\documentclass[11pt]{article}
\usepackage{graphicx,float,pict2e}
\usepackage[table]{xcolor}
\usepackage[normalem]{ulem}
\usepackage{graphicx}
\usepackage{booktabs}
\usepackage{epsfig}
\usepackage{color}
\usepackage{rotating}
\usepackage{verbatim}
\usepackage{hyperref}
\usepackage{amsmath}
\usepackage{amssymb}
\usepackage{amsfonts}
\usepackage{caption}
\usepackage{subcaption}
\title{Color-complexity enabled exhaustive color-dots identification and spatial patterns testing in images.}
\author{\normalsize Liao, Shuting\footnotemark[1], Liu, Li-Yu\footnotemark[2], Chen, Ting-An\footnotemark[2], Chen, Kuang-Yu\footnotemark[3] and Hsieh, Fushing\footnotemark[4]}
\date{}
\begin{document}
\maketitle

\begin{abstract}
Targeted color-dots with varying shapes and sizes in images are first exhaustively identified, and then their multiscale 2D geometric patterns are extracted for testing spatial uniformness in a progressive fashion. Based on color theory in physics, we develop a new color-identification algorithm relying on highly associative relations among the three color-coordinates: RGB or HSV. Such high associations critically imply low color-complexity of a color image, and renders potentials of exhaustive identification of targeted color-dots of all shapes and sizes. Via heterogeneous shaded regions and lighting conditions, our algorithm is shown being robust, practical and efficient comparing with the popular Contour and OpenCV approaches. Upon all identified color-pixels, we form color-dots as individually connected networks with shapes and sizes.  We construct minimum spanning trees (MST) as spatial geometries of dot-collectives of various size-scales. Given a size-scale, the distribution of distances between immediate neighbors in the observed MST is extracted, so do many simulated MSTs under the spatial uniformness assumption. We devise a new algorithm for testing 2D spatial uniformness based on a Hierarchical clustering tree upon all involving MSTs. Our developments are illustrated on images obtained by mimicking chemical spraying via drone in Precision Agriculture.
\end{abstract}

\footnotetext[1]{Graduate Group in Biostatistics, University of California at Davis, CA, 95616.}
\footnotetext[2]{Department of Agronomy, National Taiwan University, Taipei, Taiwan.}
\footnotetext[2]{GEOSAT Aerospace \& Technology Inc., Tainan, Taiwan}
\footnotetext[4]{Correspondence: Department of Statistics, University of California at Davis, CA, 95616. E-mail: fhsieh@ucdavis.edu.}

\section{Introduction}
Spray technologies via unmanned aerial vehicle (UAV) for liquid chemicals of fertilizers, herbicides and pesticides are at the stage of intensive researches and developments \cite{faical}. From economic and environmental perspectives, these technologies are deemed vital in Precision Agriculture \cite{mogili}. Since its wide uses will not only save costs from many aspects, particularly on human labors and illness, but also add capabilities of dynamic and optimal managements. However, the success of such technologies heavily relies of effective evaluations of their performances in terms of efficiency and precision. There might be many ways of making such evaluations. One fundamental way is to evaluate their performances by testing whether the sprayed liquid is distributed in a spatially uniform fashion upon a target area.

Recently it becomes very common that companies and research labs design their own experiments to facilitate such a fundamental testing.  One key step of this testing involves color image analysis consisting of two coupled computational tasks: exhaustive color-dots identification and spatial patterns extracting and testing. It is crucial to be able to exhaustively identify all targeted color dots of all sizes on a target area. Since each dot of sprayed chemical gives rise to two pieces of information: its amount of chemical and spatial location. Exhaustive search and extraction often are difficult to achieve computationally.  Even though color identification is a major topic in computer science, those publicly available techniques, such as Contour on gray scale and recent OpenCV with priori selected color regions, are not practical choices for dealing with heterogeneous shading on color images.

After collecting almost all colored dots and sorting out their multi-scales size-categories, two key difficulties facing us here are: the largeness of number of color-dots and geometric representations of color-dots. We need a practical unit that can embrace effectively the concept of spatial density of dot-locations. We also need a simple enough geometric representation to embed all involving units, so that structural information of spatial distribution can be extracted. Such computational endeavors for multi-scale spatial pattern extractions and testing spatial uniformness upon all size-scales are by-and-large not yet been well reported in literatures.

\begin{figure}[ht!]
\centering
\begin{subfigure}[t]{.45\textwidth}
\centering
\includegraphics[height=7cm, width=4.5cm]{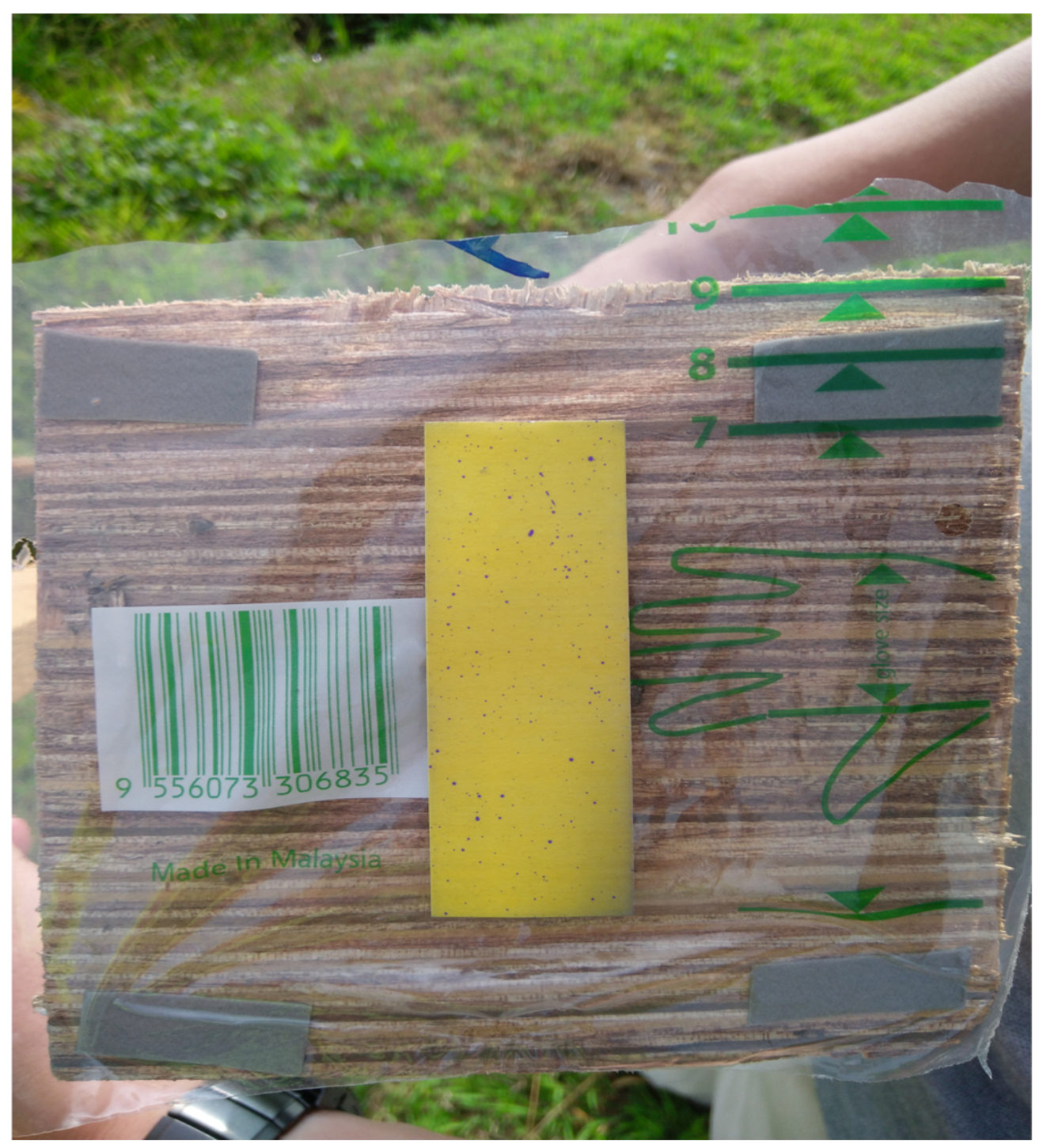}
\caption{}
\end{subfigure}
\hfill
\begin{subfigure}[t]{.45\textwidth}
\centering
\includegraphics[height=7cm, width=5.5cm]{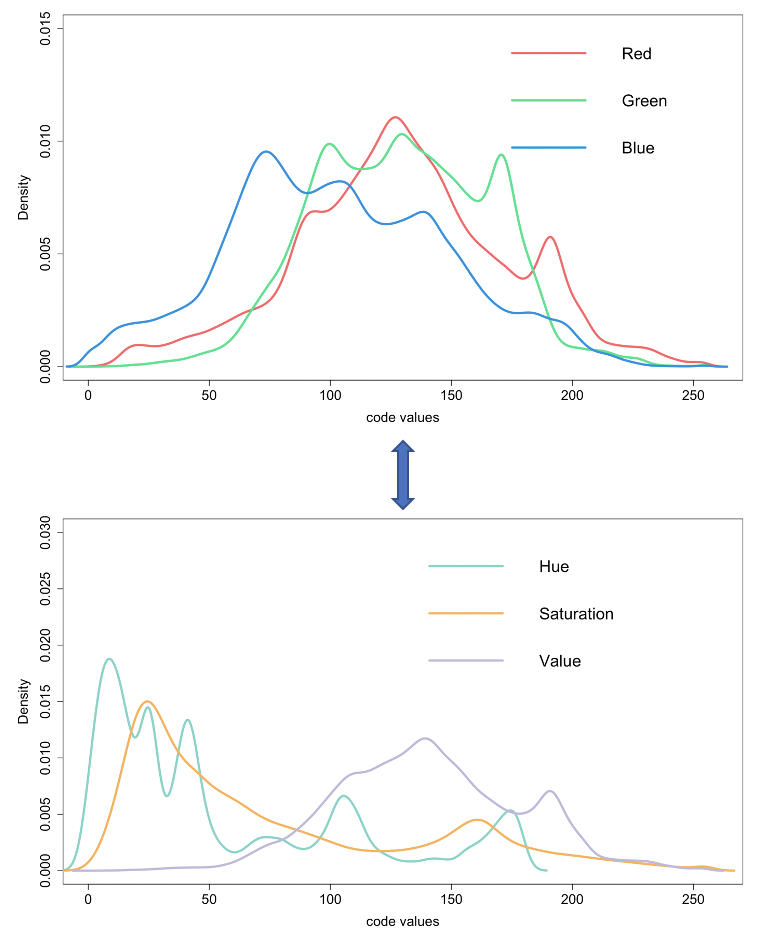}
\caption{}
\end{subfigure}
\captionsetup{justification=centering}
\caption{(a): Original image with purple dots in various sizes and shapes located on the yellow test paper; (b): RGB model and HSV model.}
\label{070420_(1-a1)}
\end{figure}

In this paper, we develop computing algorithms to resolve such coupled computational tasks. We apply our algorithms on five experimental images under rather distinct lighting conditions. We exclusively use one image for illustrating our computational developments (see Figure \ref{070420_(1-a1)}(a)).

Here we refrain from discussing Color Theory underlying the color image data, for details see \cite{schwartz}. We only briefly mention some relevant information about color data in an image like the one in Figure \ref{070420_(1-a1)}(a). As one whole, this image contains a large amount of pixels ($\sim 10^7$) including the target yellow test paper and background. Each pixel in the image has two main 3D color measurements: RGB ($[0,256)\times [0,256)\times [0,256)$) and HSV ($[0,256) \times [0,256) \times [0,256)$). These two RGB and HSV data formats, as shown in Figure \ref{070420_(1-a1)}(b), are mutual convertible. We remove most of the background and only focus on the area containing the yellow test paper, which contains around $10^6$ pixels. The image of test paper can be reconstructed as any one of RGB or HSV $10^6\times 3$ matrices.  That is, from perspectives of pixel-wise 3D intensities of RGB and HSV, a color image data here is in a form of structured data.

Based on one test paper (yellow colored-strip) and the two versions of representations of color data: RGB and HSV, our first goal is to exhaustively identify all ``purple'' dots of all sizes. Here color ``purple'' is meant to be an unspecified 3D regions within the 3D discrete space $[0,256)\times [0,256)\times [0,256)$ that reveals purple color through our visual system. Human practically has no capability to pinpoint the exact 3D region for any designated color. Since so many dots of varying sizes look like purple, it is not possible to specify a 3D color region to cover all purple dots. Therefore, the popular color-identification technique OpenCV is neither feasible nor practical for the exhaustive search purpose. Further, one test paper likely includes regions under varying shading conditions due to the photographing condition and experimental setup when the image was taken. Heterogeneous shaded images are vividly seen as well in the four images shown and analyzed in Section 5. Such existential shading will complicate choices of gray scale, and consequently reduce the efficiency of Contour approach significantly.

We propose computational algorithms to resolve this color identification issue. One physical fact of color theory plays an important role: the three dimensions of RGB or HSV are highly associated. This fact interestingly and very importantly points to that, among the $256^3$ unit cubes (of size $1\times 1\times 1$), a color image's millions of pixels typically only collectively occupy a very small number of cubes. That is, a natural color image usually has a very small ``color-complexity''. The color-complexity of the test paper in Figure \ref{070420_(23)}(a) is $0.002$. In contrast, the color-complexity of the whole image in Figure \ref{070420_(1-a1)}(a) is calculated as $\frac{393014}{256^3}\approx 0.023$. That is, the color-complexity of this test paper is only one tenth of the original image. Therefore, we indeed deal with only several thousands, not a million, of distinct colors. This is the underlying reason why our computing cost is low and our color identification is effective.

After nearly exhaustively extracting all purple pixels, we build purple dots as a connected network based on a common choice of neighborhood on 2D lattice. Then, we measure each dot's size, and classify them into several size categories. We consider two kinds of uniformness: dot-size and spatial. Upon dot-size uniformness, we aim to figure out whether the spraying machine's mechanical design is proper or not. We particularly pay attentions to behavior of the right tail of dot-size distribution (large and very large ones).

Due to the largeness of number of color-dots, we focus on spatial uniformness in the sense of density. In order to bring out sense of density, we divide the test paper into 400 squares, and categorize their densities in terms of their distributions of categorical sizes purple dots contained in them. We first propose to build a minimum spanning tree (MST) to connect all high density squares, and examine whether its tree-based distribution of distance between immediate neighboring-squares is very different from many simulated distributions derived from randomly generated  MSTs under uniform assumption. This comparison is carried out by transforming each distance distribution into a histogram with common data-driven bin-boundaries, and then collect all vector of proportions into a matrix. We build a Hierarchical Clustering (HC) tree among these distribution-IDs. Then, we develop a new algorithm to calculate p-value based on the binary structure of HC tree-geometry. We then repeatedly perform the same testing on uniformness by including less dense squares in a cumulative manner. This p-value computation is somehow novel in the sense that it is calculated based on a series of odds-ratios along a descending tree-path leading to the observed MST-tree-leaf. 

We then apply computational algorithms developed and illustrated on image no.1 to the rest of four images and report their results in Section 5.

\section{Method}
\subsection{Identification of purple pixels}
\begin{figure}[ht!]
\centering
\begin{subfigure}[t]{.4\textwidth}
\centering
\includegraphics[width=0.5\linewidth]{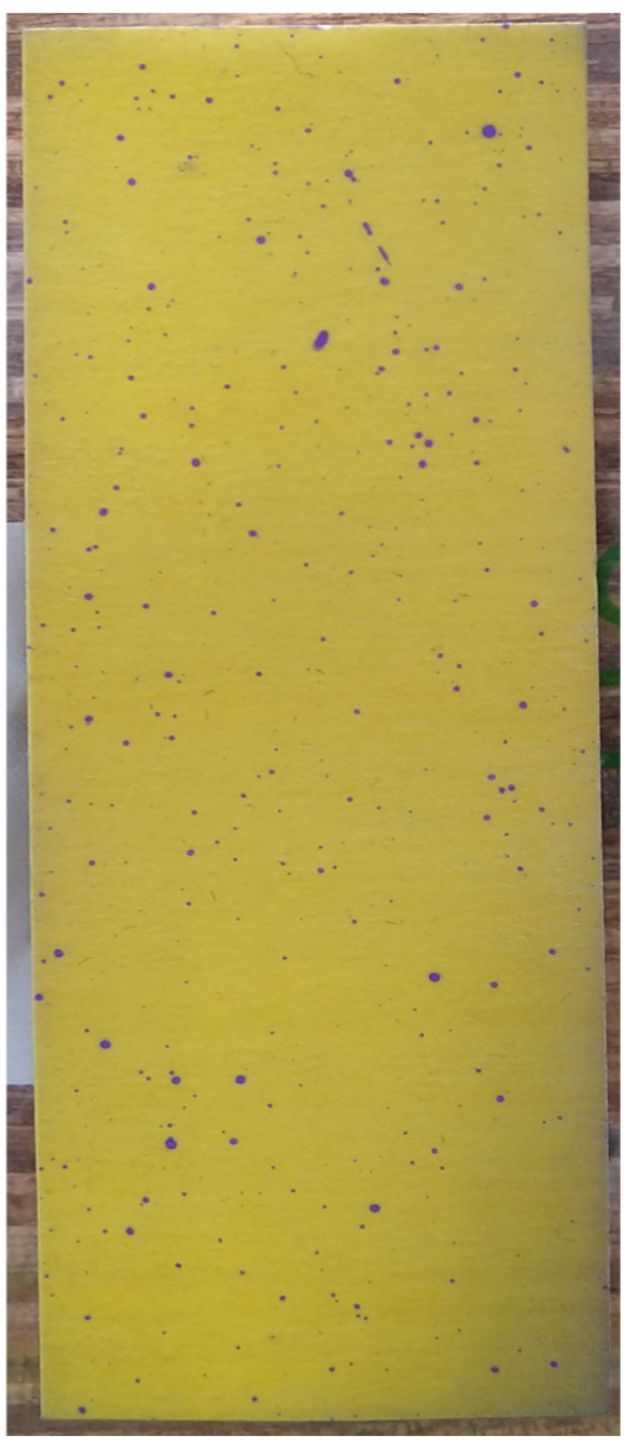}
\caption{}
\end{subfigure}
\hfill
\begin{subfigure}[t]{.4\textwidth}
\centering
\includegraphics[width=0.5\linewidth]{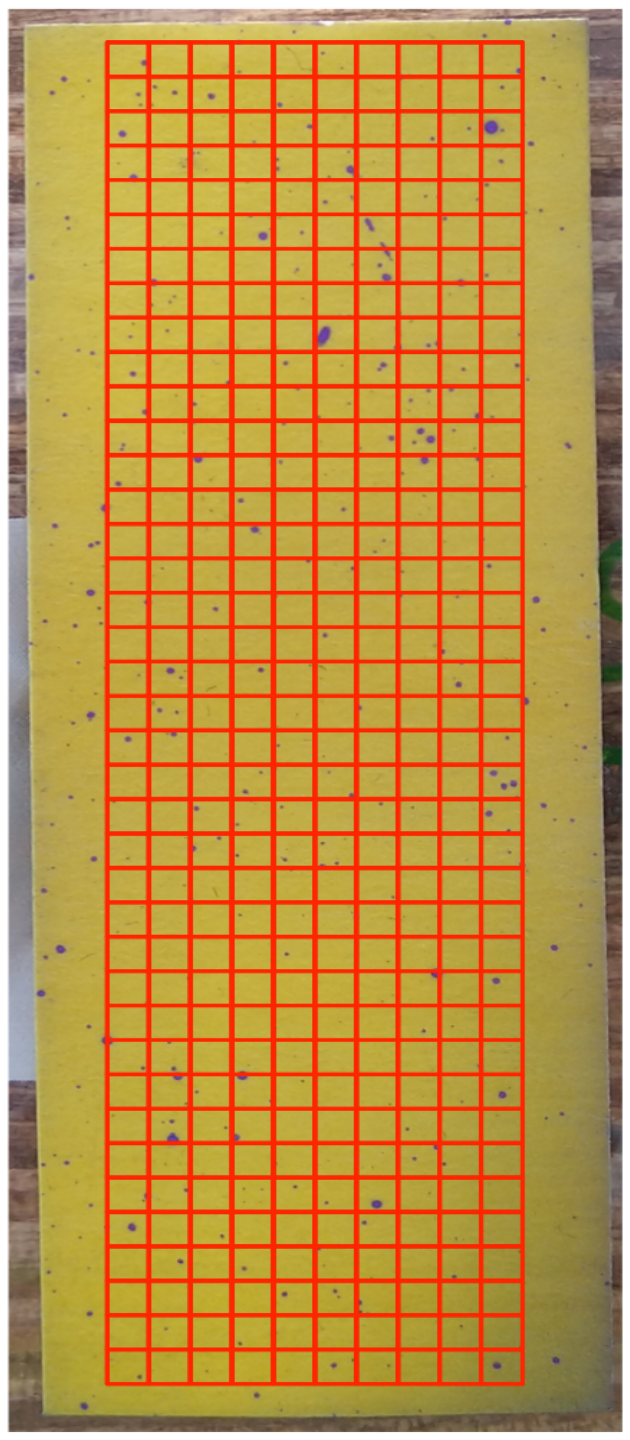}
\caption{}
\end{subfigure}
\captionsetup{justification=centering}
\caption{(a): reduced image; (b): focal area.}
\label{070420_(23)}
\end{figure}

\paragraph{}
As shown in Figure \ref{070420_(1-a1)}(a) and Figure \ref{070420_(23)}(a), it is clear that this test paper contains two main color families, yellow and purple, among the one million ($10^6$) pixels. It also evident that it contains areas of heterogeneous intensities of shades across the entire test paper. The presence of such data complications is rather common in majority of real world color images, It becomes parts of nature of data from Precision Agriculture. Since images might be taken under drastically distinct lighting conditions: with or without sun lights across different parts of days. Further, it is well known that human's visual system via brain and eye is subject to color illusions. Such illusion makes us to identify the same object with different colors under shadows as well as different backgrounds. Thus, any heterogeneously shaded image in general poses various challenges on color identification. One of the challenges is: How to do color identification in data-driven fashion? In other words, it is a necessary capability of identifying color in any image from the perspective of computer, not human.

To make computing feasible via computer, we need to have an idea how many distinct colors are indeed contained in the test paper. This is the concept referred as ``color-complexity''. Given the discrete nature of color data, it is crucial to ask: how many unit cube of $1\times 1\times 1$ among the $256^3$ ``color-unit-cubes'' are indeed occupied by the one million of color-pixels in the test paper? The answer is 28126. So the color-complexity is only $\frac{28126}{256^3} \approx 0.002$. If we enlarge the scale of unit cube to a scale of $10\times 10 \times 10$ cube, we checked and found that all potential colors contain within such a cube are still rather ``uniform'' to our raw eyesight. And, with respect to all pixels in the test paper, that there are 880 among $26\times 26 \times 26$ of such cubes being occupied. The color-complex of this test paper on this larger scale is $\frac{880}{26^3} \approx 0.05$. Hence, we decide to begin our machine learning computations upon this scale first, and go back to the unit scale afterward. It is worth emphasizing that such low color-complexity is made possible by very high non-linear associations among R\&G\&B and H\&S\&V. This is the underlying foundation to build data-driven algorithms for color identification.

Then we build a geometry among these 880 uniform color cubes. This geometry is intended to serve as a platform for our color identification. We choose this geometry to a tree for computational simplicity and practical applicability. We construct hierarchical clustering(HC) tree as follows. We use the center of mass (3D average) of pixels contained in such a cube as the cube's representative. Upon this collective of 880 representatives, the HC algorithm can work efficiently.

For completing our protocol of color identification, we take a step to tentatively avoid shady areas and background noises. Even though only involving a minority of pixels, their inclusion could yield non-negligible errors. To this end, we choose a rectangle area within the test paper as our ``focal area'', as shown in Figure \ref{070420_(23)}. This focal area is divided into 39 rows. Each row contains $2.5 \times 10^3$ pixels (Figure \ref{070420_(23)}(b)), and is further divided into 10 squares.

Our color identification begins with the following row-by-row operation. For each one row's $2.5 \times 10^3$ pixels, we identify which color-cube it belongs and then find its color-cube representative. The resultant set of distinct color-cube representatives has a size smaller than 880, surely is much smaller than $2.5 \times 10^3$. Upon this row-specific set of color-cube representatives, we apply the HC algorithm. For each row-specific HC-tree, via its bifurcation, we collect the representatives within the smaller branch as being designated as ``purple'' ones, while those in the larger branch as being ``yellow''. We further use ROC curve analysis for validation checking to avoid misclassification due to uncontrolled  environmental and lighting conditions. This validation check is performed upon each square within each row.

\begin{figure}[ht!]
\centering
\begin{subfigure}[t]{.3\textwidth}
\centering
\includegraphics[width=0.5\linewidth]{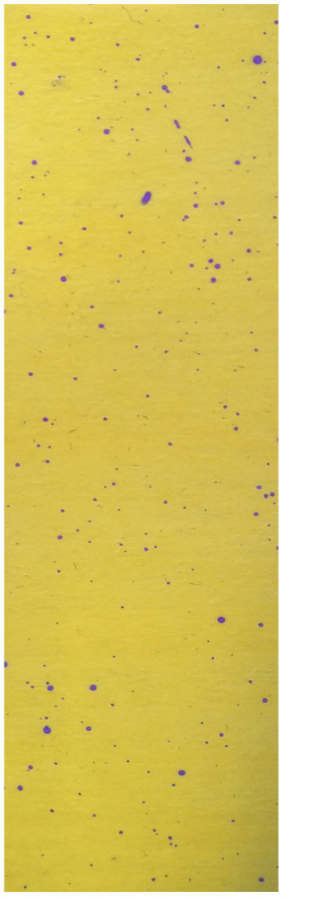}
\caption{}
\end{subfigure}
\hfill
\begin{subfigure}[t]{.3\textwidth}
\centering
\includegraphics[width=0.512\linewidth]{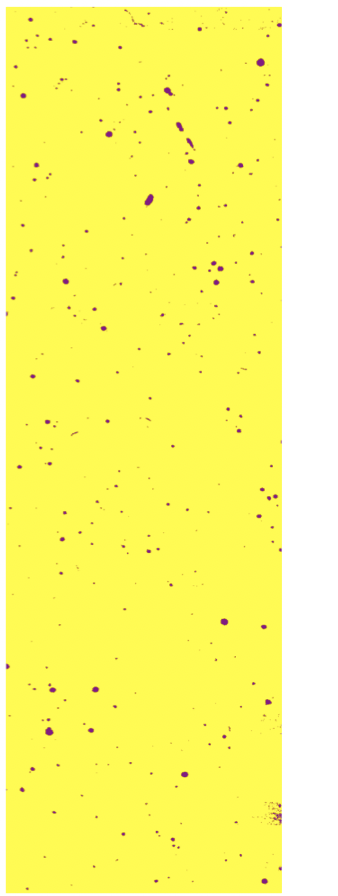}
\caption{}
\end{subfigure}
\hfill
\begin{subfigure}[t]{.3\textwidth}
\centering
\includegraphics[width=0.5\linewidth]{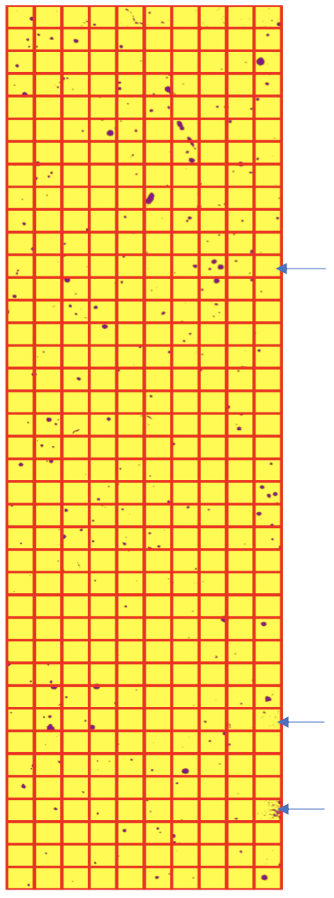}
\caption{}
\end{subfigure}
\captionsetup{justification=centering}
\caption{(a): original focal area; (b): predicted focal area by RGB file; (c) validation check indicating three squares need "cleaning".}
\label{071120_(123)}
\end{figure}

The ensemble of color-identification on the focal area via RGB data file is shown in Figure \ref{071120_(123)} together with results of square-by-square validation check. There are three squares have obvious misclassification. We ``clean'' these three squares by assigning all pixels in these three squares into the yellow group.

\subsection{recovering the whole yellow test paper}
Given that pixels outside of the focal area have higher potential being subject to shade or other noises, their color identification need extra efforts. We propose the following remedy based on our experiences derived from our explorations and experiments. Upon RGB data format, we need to employ $1 \times 1 \times 1$ small RGB color-cubes, denoted as the scale of ``$n=1$''. That is, we need to drastically sharpen the color-uniformness within each color-cube. So we have to pay more computing cost to achieve the goal of color identification with RGB data, even though, we still enjoy reduction of color complexity because only $0.2\%$ of $1 \times 1 \times 1$  RGB unit color-cubes are occupied.

Consequently, we collect the centers of all color-cubes, which have ever being occupied by an identified purple pixel in the focal area. And likewise collect the centers of all color-cubes, which have ever being occupied by an identified yellow pixel in the focal area. Among the pool of these two collections of centers of color-cubes, we compute a closest neighbor to each pixel outside of the focal area, and then declare the color-identification accordingly. In this way, we are able to capture the majority of purple pixels and avoid misclassification as much as possible.

\begin{figure}[ht!]
\centering
\begin{subfigure}[t]{.2\textwidth}
\centering
\includegraphics[width=0.9\linewidth]{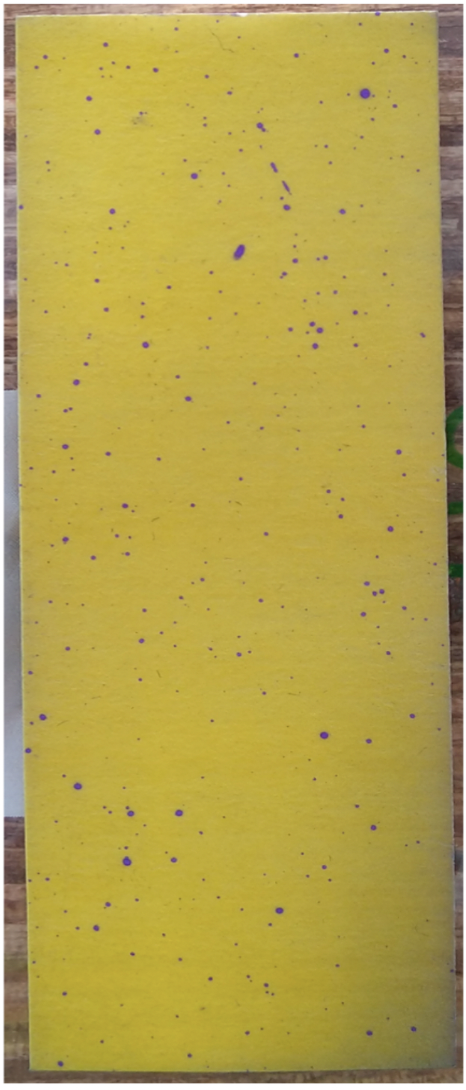}
\caption{}
\end{subfigure}
\hfill
\begin{subfigure}[t]{.2\textwidth}
\centering
\includegraphics[width=0.9\linewidth]{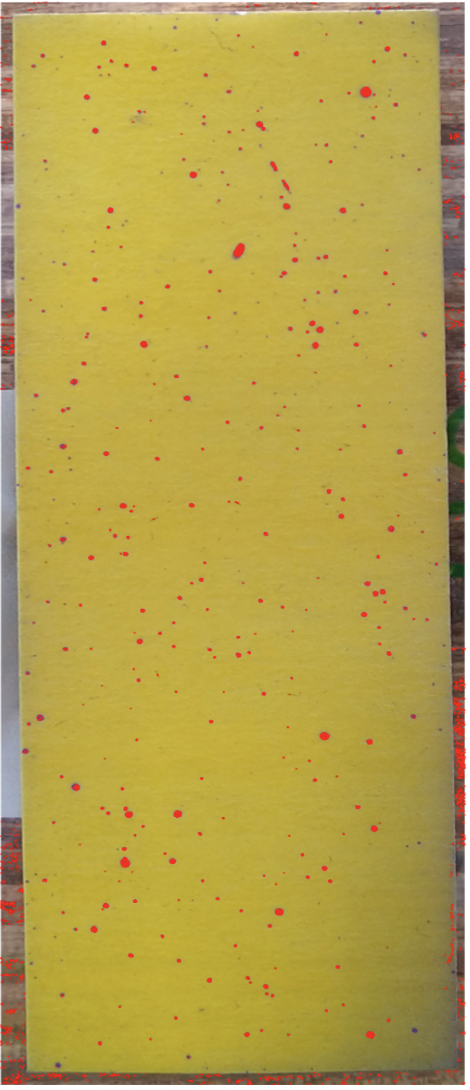}
\caption{}
\end{subfigure}
\hfill
\begin{subfigure}[t]{.2\textwidth}
\centering
\includegraphics[width=0.9\linewidth]{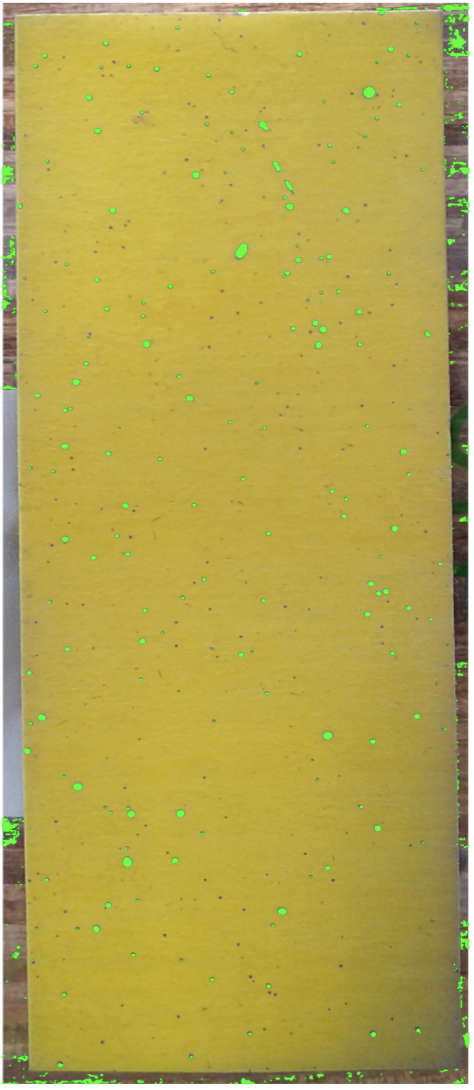}
\caption{}
\end{subfigure}
\hfill
\begin{subfigure}[t]{.2\textwidth}
\centering
\includegraphics[width=0.9\linewidth]{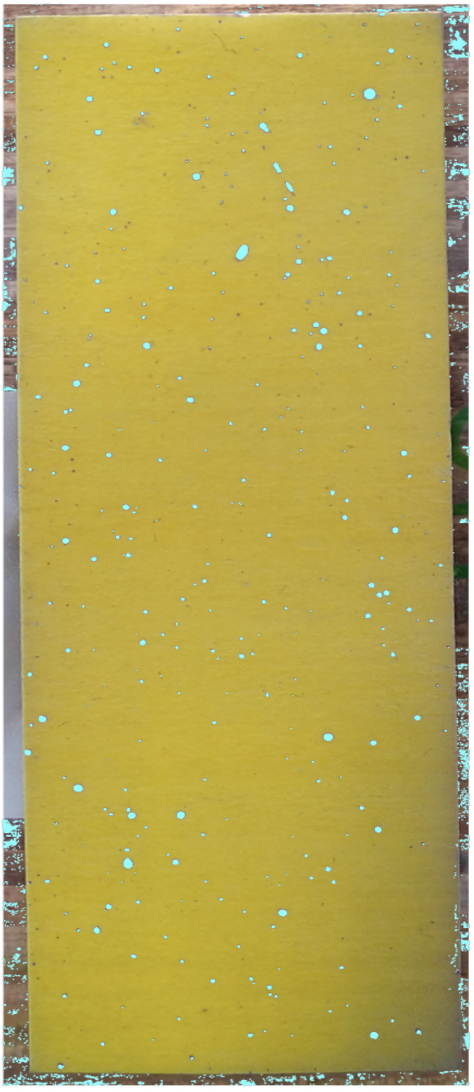}
\caption{}
\end{subfigure}
\captionsetup{justification=centering}
\caption{(a): original reduced image; (b): recovering (in red) by RGB file ($n=1$); (c): recovering (in green) by HSV file ($n=10$); (d): recovering (in light blue) by combining RGB file ($n=1$) and HSV file ($n=10$).}
\label{071120_(4567)}
\end{figure}

As for HSV data format, we still employ the scale $n=10$, i.e. $10 \times 10 \times 10$ HSV color-cubes. We obtain the recovering by both RGB file and HSV file separately. It turns out that RGB file helps identify more pixels in smaller purple dots, while HSV file helps identify more pixels near the bottom and top where the RGB file fails. The two results suggest a better recovering scheme as simply combining these two results together. All results are shown in Figure \ref{071120_(4567)}.

\section{Testing uniformness via sizes}
As it is intuitively known that a spraying device typically mixes air with liquid, and then push the mixture out. The mixing of air and liquid is determined by a set of tuning parameters. Mechanically speaking, different sets of tuning parameters surely give rise to distinct degrees of inhomogeneous mixing. Consequently, the droplets out of the device are likely heterogeneous in size.  So, some tuning parameters are better than others. One merit of exhaustive identification of targeted color-dots contained in an image is to check validity of a parameter-setup of spraying device. For this merit, there are two natural measure sizes of a droplet, which is an identified connected purple-pixel. The first measure is to count number of connecting pixels. The second one is the radius of the smallest circle containing all connecting pixels. Accordingly, the best set of tuning parameters should ideally produce the Poisson distribution with respect to the counting measure, and an Exponential with respect to the continuous measure.

We consider the target collection of color-dots identified via the approach of combining the RGB and HSV data, see Figure \ref{071120_(4567)}(d). We first compute the MLEs of intensity parameters, $\lambda_P$ and $\lambda_E$, under the Poisson and Exponential distribution assumptions, respectively, based on the two data sets derived from the target collection of purple-dots within the test paper.

\begin{figure}[ht!]
\centering
\begin{subfigure}[t]{.45\textwidth}
\centering
\includegraphics[height=4.5cm]{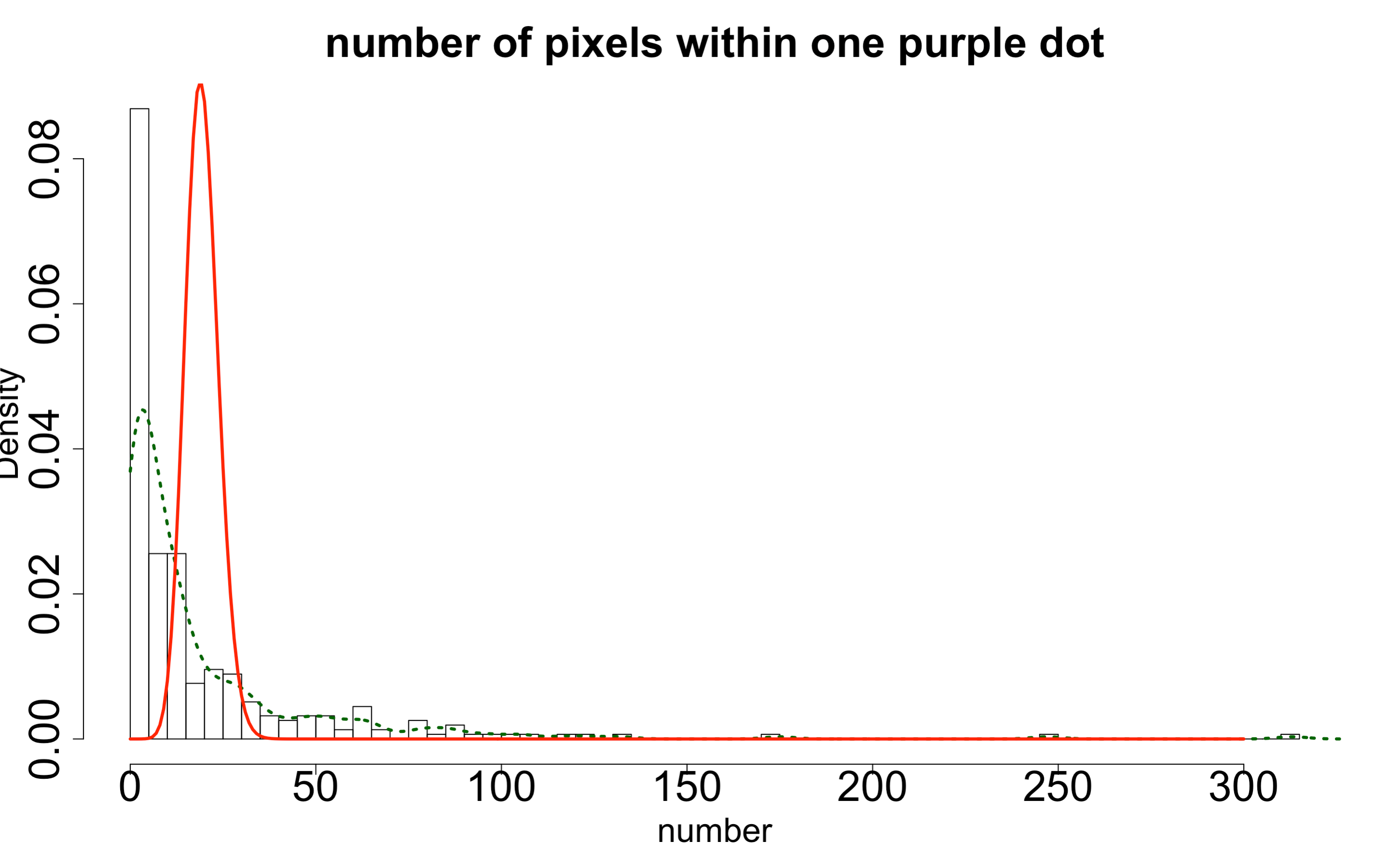}
\caption{}
\end{subfigure}
\hfill
\begin{subfigure}[t]{.45\textwidth}
\centering
  \includegraphics[height=4.5cm]{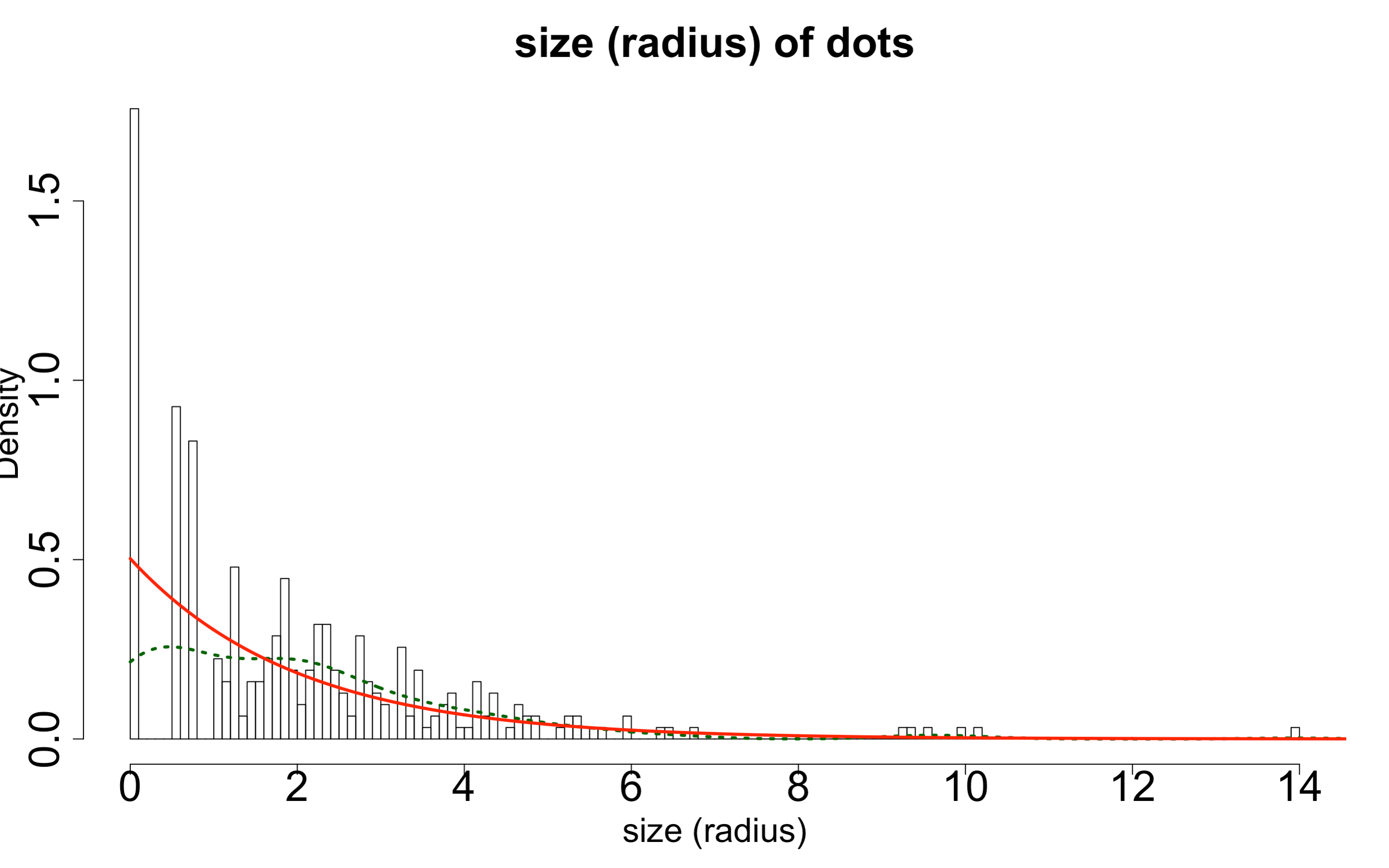}\llap{\raisebox{1.3cm}{\includegraphics[height=2.5cm]{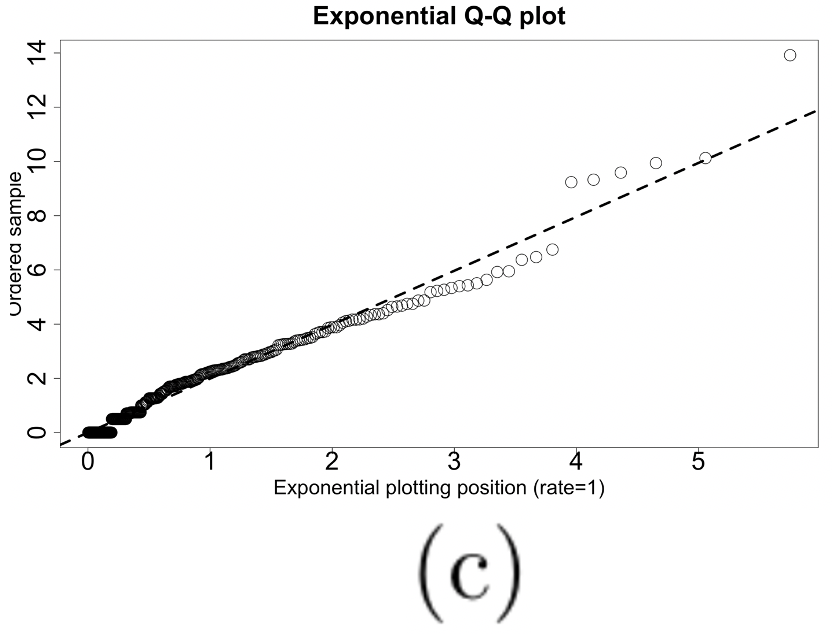}}}
\caption{}
\end{subfigure}
\captionsetup{justification=centering}
\caption{(a): The histogram of pixel-counts of identified purple-dots superimposed with Poisson distribution specified MLE of $\lambda_P$(red curve) and kernel density estimates (green dash curve);(b): The histogram of dot-sizes of identified purple-dots superimposed with Exponential distribution specified MLE of $\lambda_E$ (red curve) and kernel density estimates (green dash curve);(c): Empirical Q-Q plot (Circle curve) vs Exponential Q-Q plot specified by MLE of $\lambda_E$(dash line).}
\label{071220_(125)}
\end{figure}

Based on the pixel-count data set, the Poisson distribution specified by MLE of $\lambda_P$ is computed and superimposed onto the histogram constructed based on pixel-counts from the target collection of purple dots, as shown in Figure \ref{071220_(125)}(a). It is evident that many identified purple-dots have large pixel counts that can not be accounted for by Poisson distribution. We can draw a similar conclusion based on the dot-size distribution with superimposed Exponential distribution specified by MLE of $\lambda_E$ shown in Figure \ref{071220_(125)}(b).

While the Q-Q plot Exponential distribution specified by MLE of $\lambda_E$ is compared with empirical Q-Q plot of continuous purple-dots' sizes, as shown in Figure \ref{071220_(125)}(c). We see evident departures from this Q-Q plots comparison. Further, we run the Kolmogorov-Smirnov test, which suggests the observed dot-size is not following an Exponential distribution (with p-value $< 0.05$)

\section{Testing spatial uniformness via rectangle neighborhood}
In this section, we construct our major algorithmic developments for testing against the 2D spatial uniformness. We adopt the concept of 2D neighborhood into 2D spatial characteristics. The reasons behind are that the number of identified purple-dots are too many, and their sizes are rather heterogeneous. This neighborhood concept directly links to the idea of spatial density, which is a proper expression for addressing spatial uniformness here.

Given that we specifically divide the entire target area into 400 small rectangles, one rectangle is taken as one 2D neighborhood. On this collective of rectangles, we pretend as if they are uniformly colored with an intensity (density) of purple depending on all purple-dots contained in it. In this fashion, we consider the 2D spatial uniformness among 2D-entities of 400 rectangles. Since all identified purple-dots have been classified into three categories of sizes: small, medium and large. So an intensity of purple dots in a rectangle would be also categorized, as given below. We apply Hierarchical Clustering (HC) algorithm to guide this categorization. The categorizing protocol is devised as follows.

\begin{figure}[ht!]
\centering
\begin{subfigure}[t]{.35\textwidth}
\centering
\includegraphics[width=0.45\linewidth]{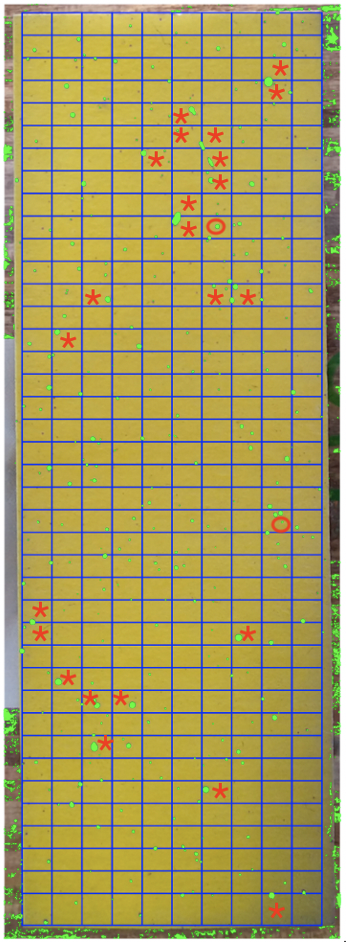}
\caption{}
\end{subfigure}
\hfill
\begin{subfigure}[t]{.5\textwidth}
\centering
\includegraphics[height=7cm, width=8cm]{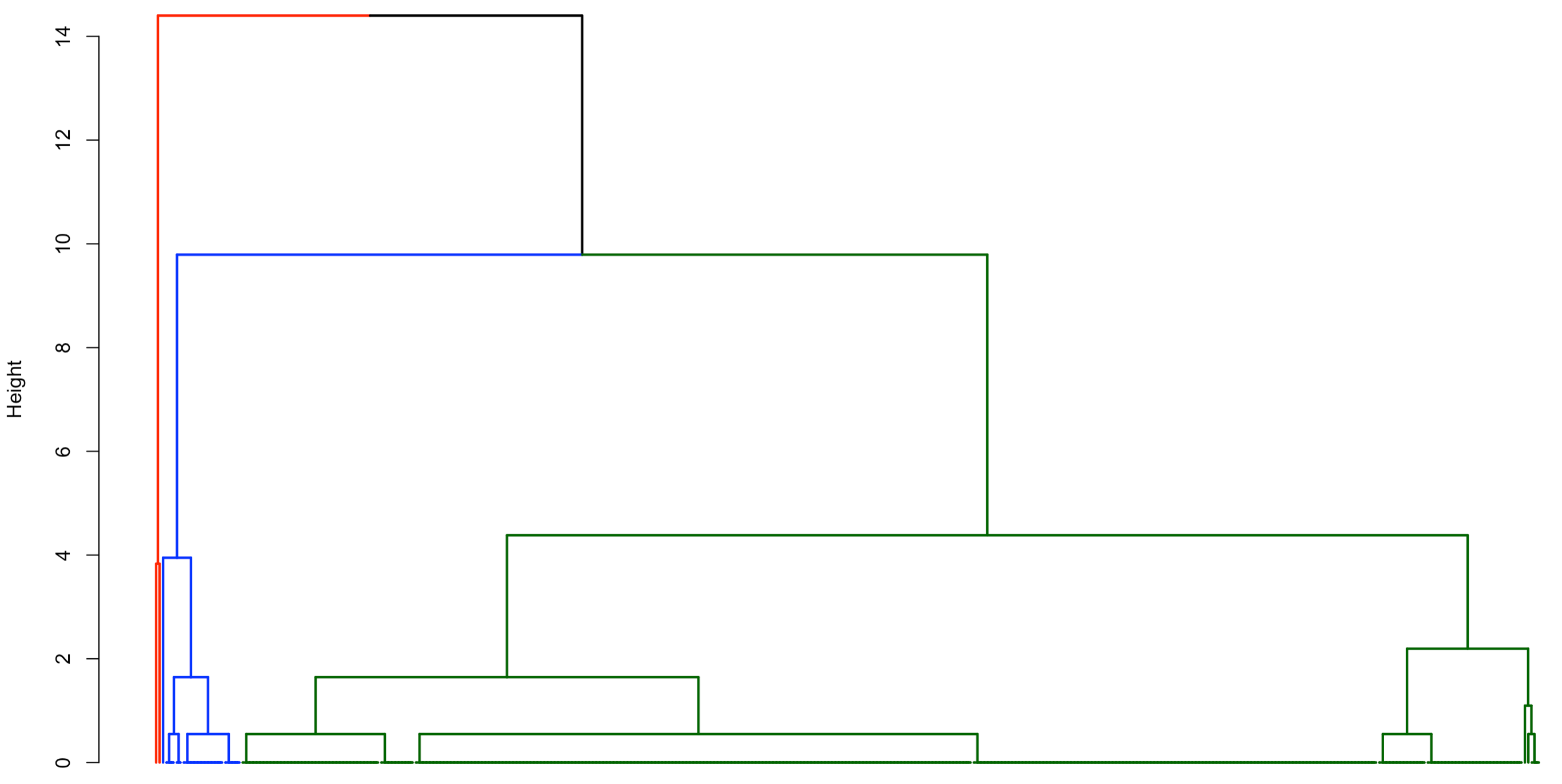}
\caption{}
\end{subfigure}
\captionsetup{justification=centering}
\caption{(a) \textcolor{red}{$\star$}: rectangles with $\geq$ 1 big dot are identified (in blue branch in (b)); \textcolor{red}{$o$}: rectangles with $\geq$ 2 medium dots (in red branch in (b)); (b) HC tree of 400 rectangles with 3 indicated clusters.}
\label{071120_(21-22)}
\end{figure}

We count the numbers of small, medium and large purple-dots contained in a rectangle as the 3 features for this rectangle. That is, each rectangle of many pixels as a unstructured data format is characterized by a 3-dim vector of counts. Via this characterization, we transform a rectangle into a structured data format. We employ a distance measure that is a weighted version of Euclidean distance in $R^3$. To reflect larger dot-size giving rise to higher purple-color intensity, this weighting scheme is specified with respect to the 3 averaged sizes: small, medium and large, of purple-dots. With this weighted distance measure, we build a $400 \times 400$ distance matrix. A HC-tree is computed and reported in Figure \ref{071120_(21-22)}(b).

Upon this HC-tree, we can see two small branches (red and blue colored) constituting a clear pattern: their member rectangles either contain at least one large or two medium dots. Locations of these rectangles are shown in Figure \ref{071120_(21-22)}(a). This data-driven pattern leads us to explore the intensity spectrum via hierarchical clustering tree on these 400 rectangles.

\begin{figure}[ht!]
\centering
\begin{subfigure}[t]{.3\textwidth}
\centering
\includegraphics[width=1\linewidth]{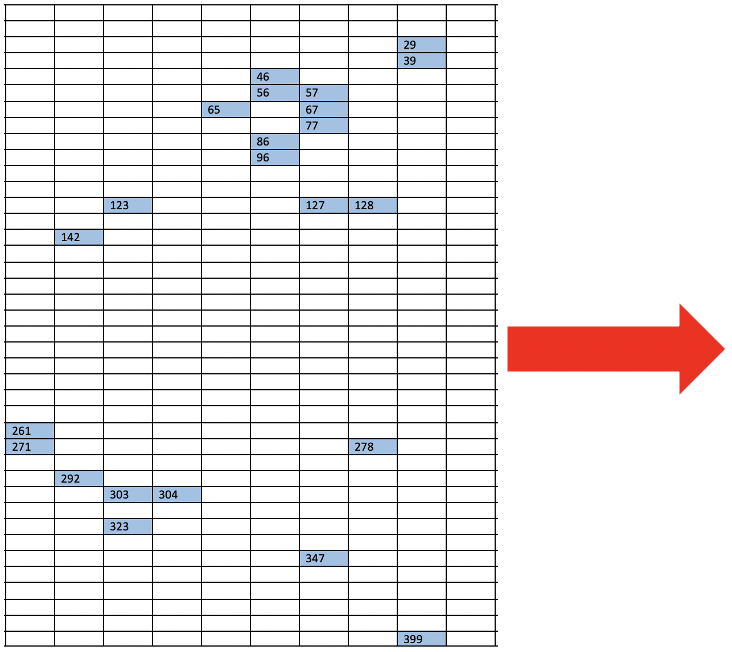}
\caption{}
\end{subfigure}
\hfill
\begin{subfigure}[t]{.3\textwidth}
\centering
\includegraphics[width=1\linewidth]{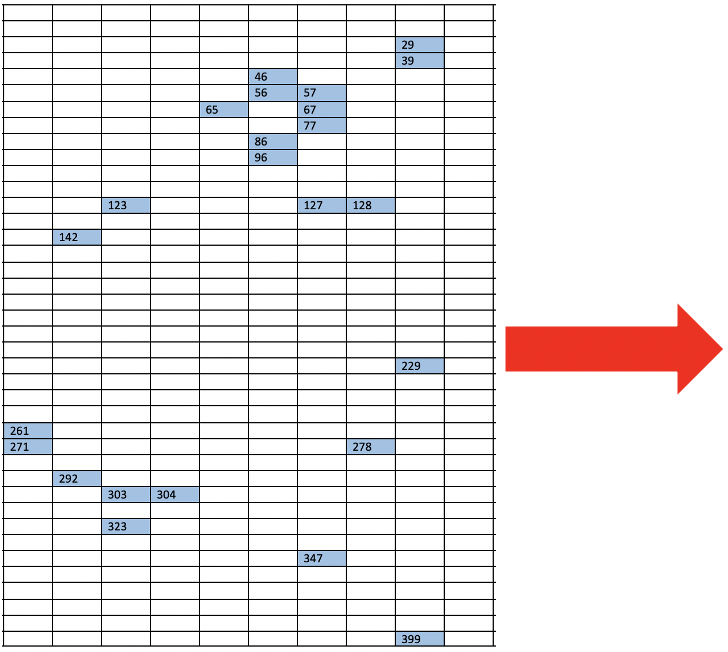}
\caption{}
\end{subfigure}
\hfill
\begin{subfigure}[t]{.3\textwidth}
\centering
\includegraphics[width=1\linewidth]{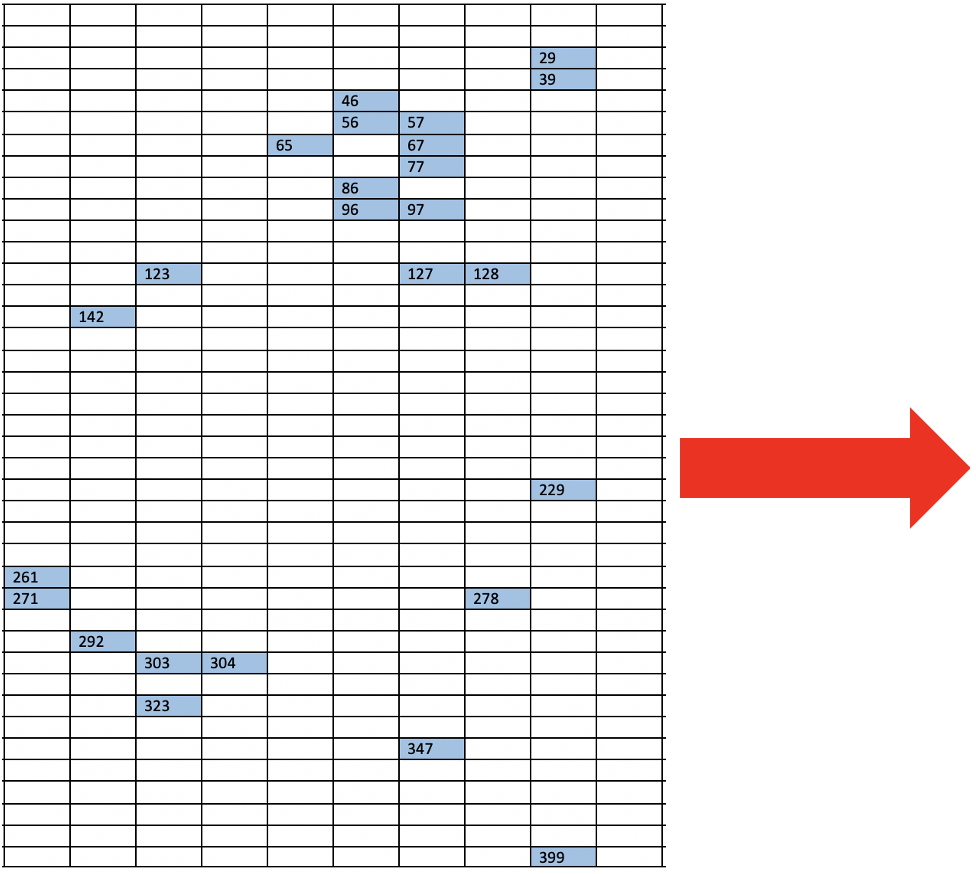}
\caption{}
\end{subfigure}
\vspace{1cm}
\begin{subfigure}[t]{.3\textwidth}
\centering
\includegraphics[width=1\linewidth]{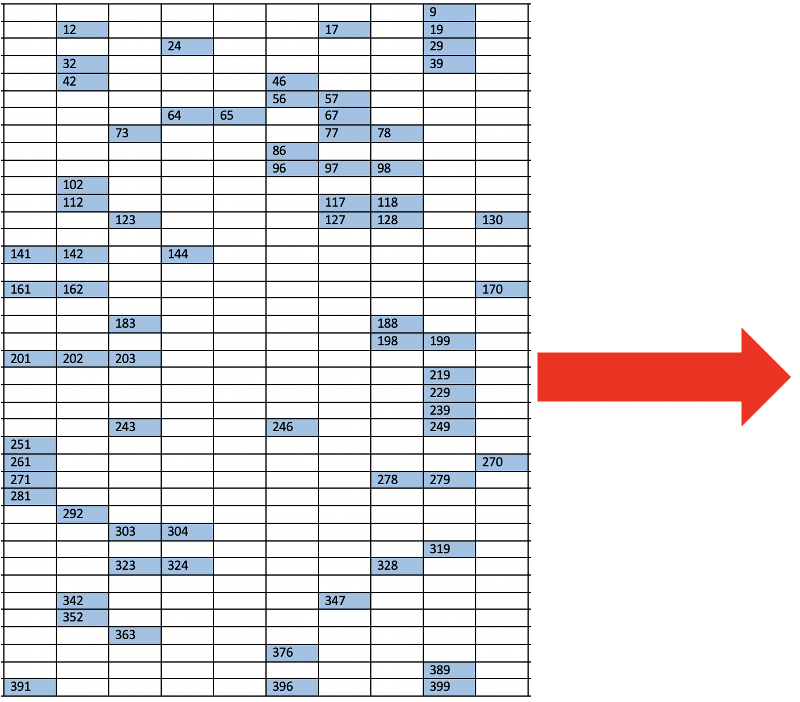}
\caption{}
\end{subfigure}
\begin{subfigure}[t]{.3\textwidth}
\centering
\includegraphics[width=1\linewidth]{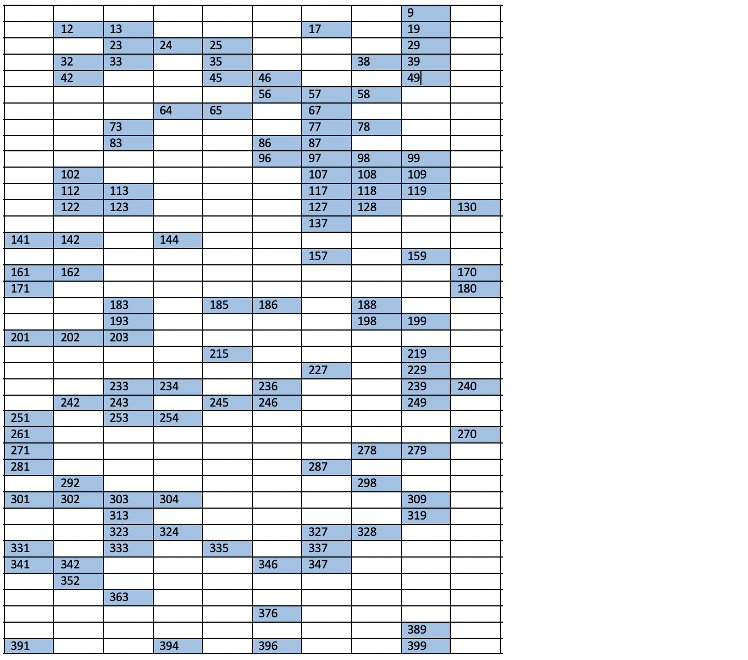}
\caption{}
\end{subfigure}
\captionsetup{justification=centering}
\caption{The flow of rectangles' 2D -distribution, from large dots to medium and small dots; some specific spatial patterns are present; (a): rectangles with 1 large dot; (b): add rectangles with $\geq$ 3 medium dots; (c): add rectangles with $\geq$ 2 medium dots; (d): add rectangles with $\geq$ 1 medium dot; (e): add rectangles with $\geq$2 small dots; the blank are with only 1 or 0 small dot.}
\label{071120_(8-12)}
\end{figure}

More or less based on this HC tree, we further qualitatively determine four categories of density within a rectangle as follows:  A rectangle contains
\begin{description}
\item[R1:] [Highly-dense] one or more large dots, or two or more medium  dots;
\item[R2:] [Dense] one medium dot;
\item[R3:] [Sparse] 2 or more small dots;
\item[R4:] [Extremely-Sparse] only one small dot or empty.
\end{description}
We found that there are 25 rectangles belonging to the Highly-dense category, which are located on the blue and red branches of the HC tree in Figure \ref{071120_(21-22)}(b). The spatial geometries of these four categories of rectangles can be seen in Figure \ref{071120_(8-12)} in a cumulative manner.

\begin{figure}[ht!]
\centering
\begin{subfigure}[t]{.4\textwidth}
\centering
\includegraphics[width=0.8\linewidth]{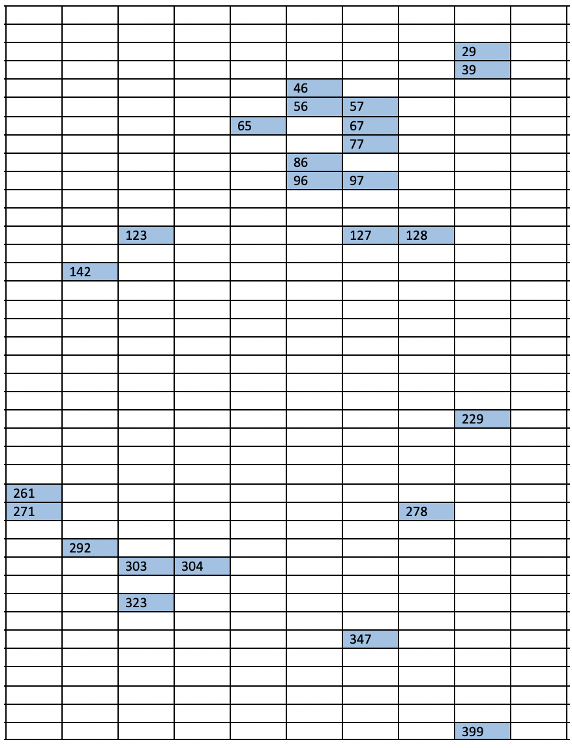}
\caption{}
\end{subfigure}
\hfill
\begin{subfigure}[t]{.4\textwidth}
\centering
\includegraphics[width=1\linewidth]{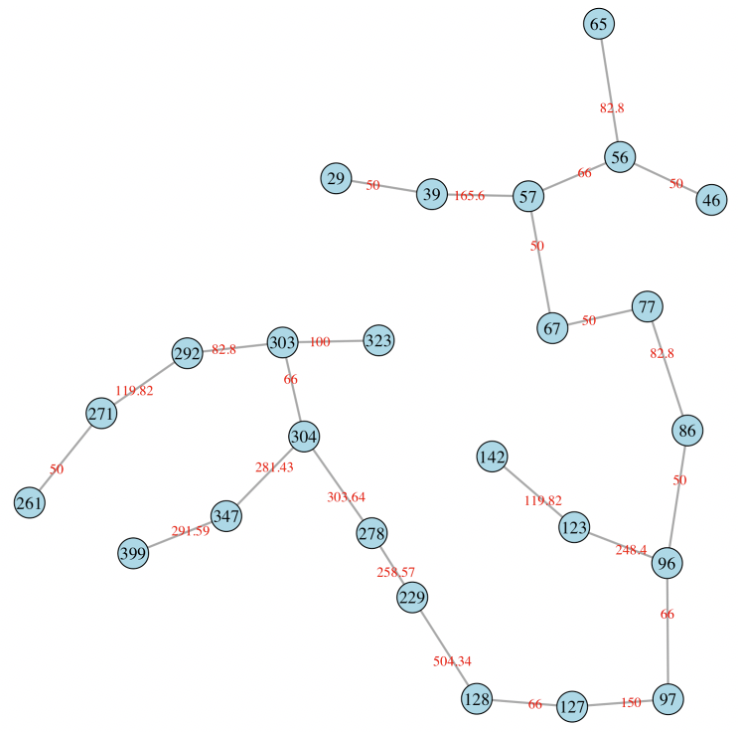}
\caption{}
\end{subfigure}
\captionsetup{justification=centering}
\caption{(a): interested 2D distribution of rectangles with $\geq$ 2 medium dots; (b): using the corresponding Minimum Spanning Tree (MST) to capture the spatial pattern.}
\label{071120_(13-14)}
\end{figure}

\paragraph{Via Minimum Spanning Tree (MST)}
Upon these 25 highly-dense rectangles, we construct a Minimum Spanning Tree (MST), denoted as $M^{obs}$, as shown in Figure \ref{071120_(13-14)}(b). The intuitive idea underlying MST is that its tree geometry, which spans a subregion by having one tree-leaf linking to one of its close neighbors, will reflect possibly heterogeneous degrees of spatial concentration among the 25 rectangle members. One way of expressing such heterogeneity in spatial concentration of a MST is to look through the empirical distribution (or histogram) of distances among all connected immediate-tree-neighbors. Such an empirical distribution (or histogram) is an informative summarizing exhibition for the degree of heterogeneous concentration pertaining to a MST. We particularly look out for the extremely high concentrations, which will lead a MST's empirical distribution of distance, or its histogram, to reveal a single mode located at a small distance value.

With aforementioned focal characteristics in mind, to test whether $M^{obs}$ is coherent with the 2D spatial uniformness hypothesis, we compare $M^{obs}$'s empirical distribution (or histogram) with $B(=500)$ randomly generated MSTs' empirical distributions (or histograms). 25 numbers are sampled randomly from a collection of digits $\{1, 2,..., 400\}$ with equal probability. We repeat this simulation scheme for $B(=500)$ times with independency. We accordingly generate corresponding $B$ MSTs, denoted as $\{ M_b \}^B_{b=1}$. So we have $B$ simulated empirical distributions (or histograms) under the spatial uniformness hypotheses. To compare $M^{obs}$ with $\{ M_b \}^B_{b=1}$ via their empirical distributions (or histograms), we propose two approaches: Mann-Whitney statistics and unsupervised machine learning approach.

\begin{figure}[ht!]
\centering
\begin{subfigure}[t]{.45\textwidth}
\centering
\includegraphics[width=1\linewidth]{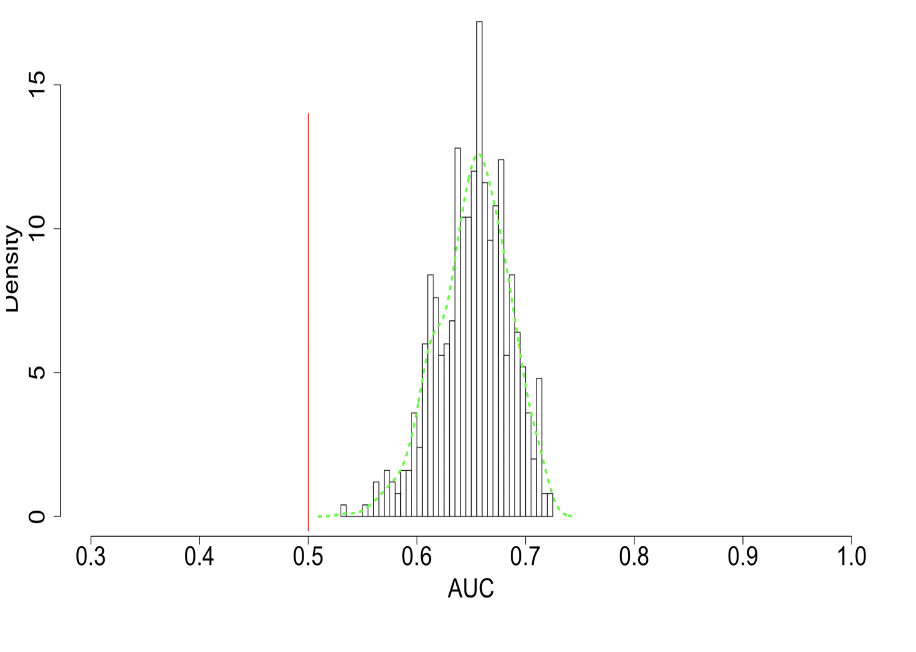}
\caption{}
\end{subfigure}
\hfill
\begin{subfigure}[t]{.45\textwidth}
\centering
\includegraphics[width=1\linewidth]{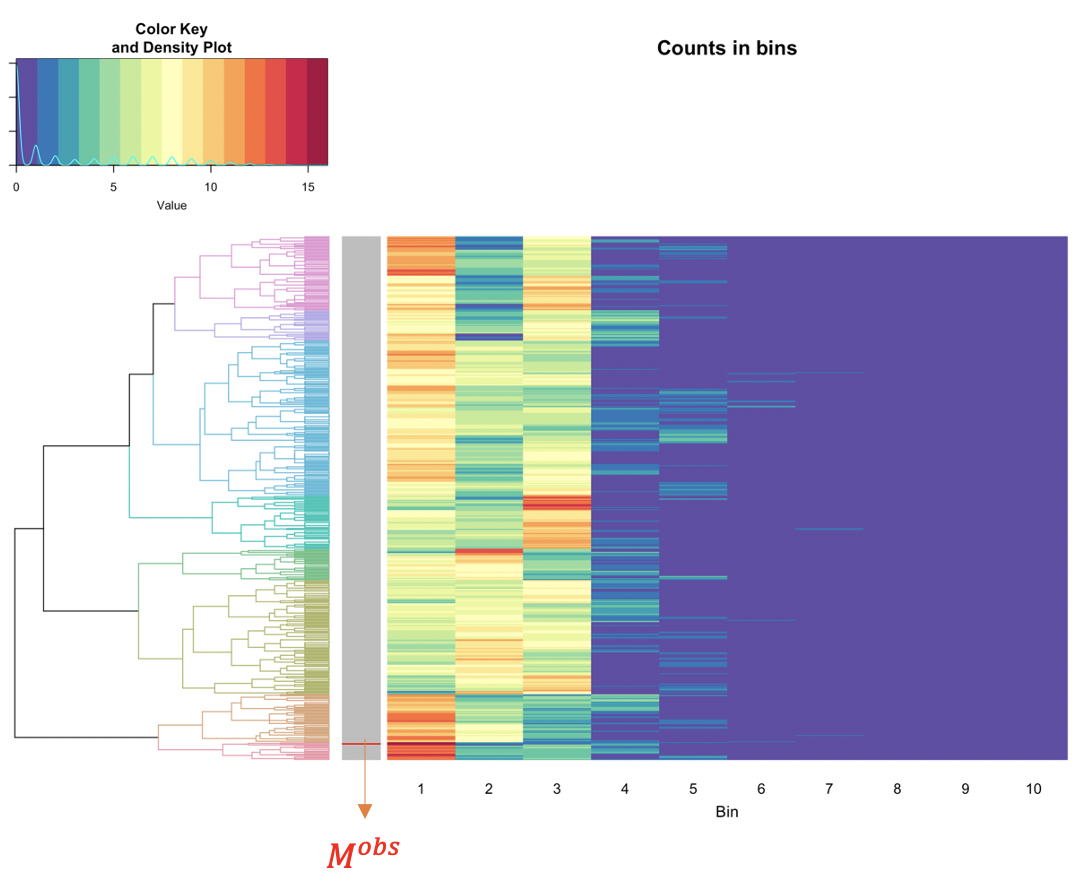}
\caption{}
\end{subfigure}
\captionsetup{justification=centering}
\caption{(a): ROC curve anlysis results; (b): HC algorithm with heatmap, product of odds ($PO=0.023$) and p-value $p(M^{obs})=0.002$.}
\label{071120_(15-16)}
\end{figure}

\paragraph{Mann-Whitney statistics:} The Mann-Whitney statistics compares a pair of distributions, say $F(.)$ and $G(.)$, via the building an Operation Characteristic Curve (ROC): $1-G(F^{-1}(1-w))$ with $F^{-1}(w)$ the quantile function of $F(.)$. So we compute $B$ ROC curves for the $B$ pairs empirical distributions pertaining to $B$ pairs of $(M^{obs}, M_b)$ with $b=1,..., B$. So $B$ pieces of area-under-curve (AUC) are calculated. The histogram of $B$ AUC values are shown in Figure \ref{071120_(15-16)}(a). The marked 0.5 value of AUC, which corresponds to the case of $F(.)=G(.)$, is seen being far away from this histogram. This fact strongly indicates that the $M^{obs}$'s empirical distribution (or histogram) is stochastically smaller than the empirical distributions (or histograms) of $\{ M_b \}^B_{b=1}$ in a persistent manner. It is known that Mann-Whitney statistics is an efficient U-statistics when both distributions have the same distributional shape. There is no guarantee for being this case throughout. In fact, despite the required assumption, its fundamental drawback rests on the fact that one dimensional statistics is unlikely to reveal structural differences between two distributions because of their high dimensionality. That is why unsupervised machine learning approach is needed.

\paragraph{Unsupervised machine learning approach:} We want to literally compare these $B+1$ empirical distributions derived from $M^{obs}$ with $\{ M_b \}^B_{b=1}$. To facilitate such a direct comparison, we pool together all distance values from these $B+1$ empirical distributions, and then build a histogram with 10-bins. With such data-driven bin boundaries, we transform each empirical distribution into a 10-dim vector of counts. These $B+1$ 10-dim vectors are arranged along the row-axis a $(B+1)\times 10 $ matrix.

Due to the equal total counts on all rows, we simply adopt Euclidean distance and then calculate a $(B+1)\times (B+1)$, with which we
apply the Hierarchical Clustering (HC) algorithm to build a HC tree, denoted as ${\cal T}$, and superimpose it onto the row-axis of $(B+1)\times 10 $ matrix, as shown in Figure \ref{071120_(15-16)}(b). The tree-leaf of $M^{obs}$ is marked onto the HC tree upon this rearranged matrix, which is called a heatmap.

The resultant heatmap explicitly shows why $M^{obs}$ is found among an extreme subgroup of the ensemble $\{ M_b \}^B_{b=1}$. The visible pattern is that the 25 members of $M^{obs}$ have dominantly many extremely small distances among immediate neighbors. This pattern indeed indicates high degrees of concentration among 25 members of $M^{obs}$. This is a strong piece of evidence against the spatial uniformness assumption. How strong it is? Next we develop an algorithm to do such an evaluation.

The HC tree ${\cal T}$ is binary. Therefore each of $B+1$ tree-leave can be located by a binary tree-descending tracing process. If we adopt a coding scheme to encode the left-branching with a code-$0$ and right-branching with a code-$1$ at each internal node of ${\cal T}$. Then each tree-leaf is encode by a binary code sequence. Denote the binary code sequence for $M^{obs}$ as $ <d^o_1, d^o_2,...d^o_{K_o}>$ and a code sequence for $M_b$ as $ <d^b_1, d^b_2,...d^b_{K_b}>$ with $b=1,.., B$. The coding lengths $K_o$ and $\{K_b\}^B_{b=1}$ vary from one tree-leaf to another tree-leaf.

\begin{figure}[ht!]
    \centering
    \includegraphics[width=0.9\textwidth]{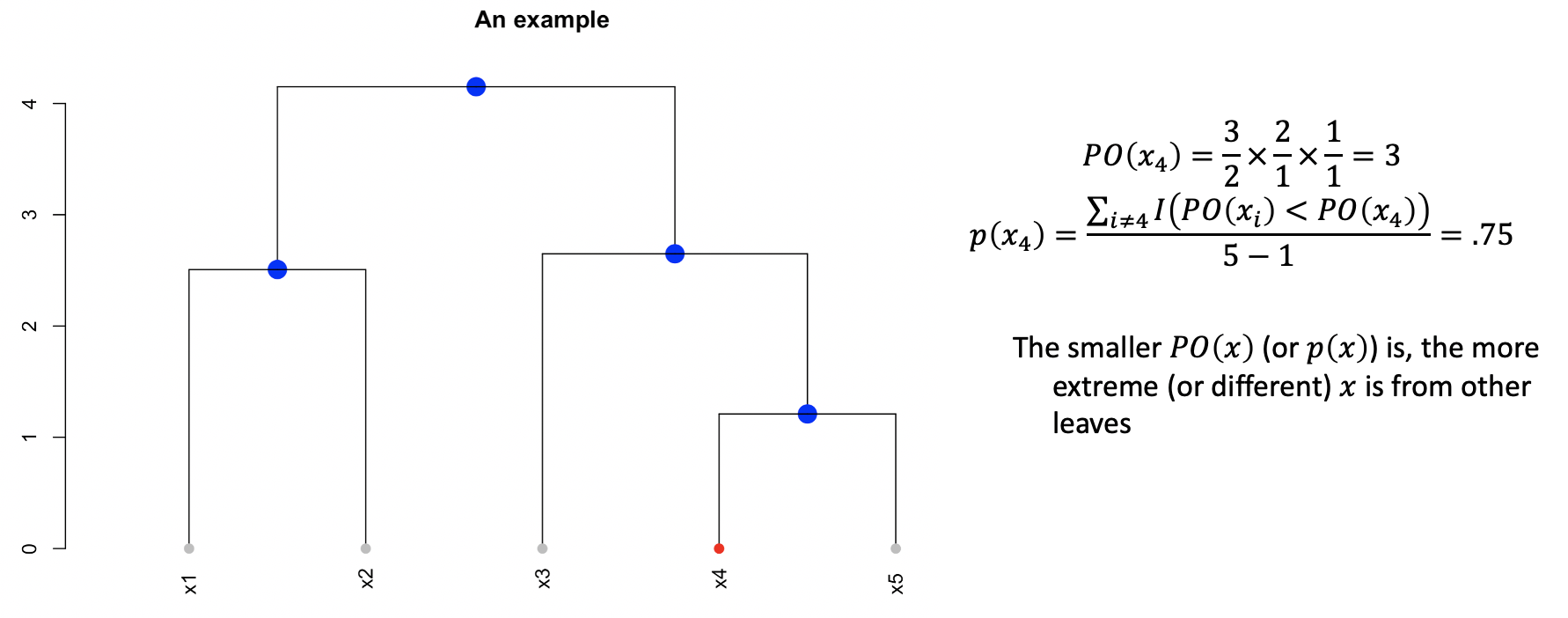}
    \captionsetup{justification=centering}
    \caption{An example as illustration of product of odds ($PO$) and p-value.}
        \label{071220_6}
\end{figure}

Further, with the binary code sequence as the descending path of bifurcating for locating $M^{obs}$, we denote the left and right branches at $k$-th bifurcation as $L_{d^o_k}$ and $R_{d^o_k}$ with $k=1,..., K_o$. The sizes of the two branches are denoted as $|L_{d^o_k}|$ and $|R_{d^o_k}|$. Then the size of the branch containing $M^{obs}$ at $k$-th bifurcation is calculated as
\[
|L_{d^o_k}|^{(1-d^o_k)}|R_{d^o_k}|^{(d^o_k)}.
\]
Then there is an odds of correctly guessing which branch contains $M^{obs}$ is calculated as:
\[
PO[d^o_k|M^{obs}]=\frac{|L_{d^o_k}|^{(1-d^o_k)}|R_{d^o_k}|^{(d^o_k)}}{|L_{d^o_k}|^{(d^o_k)}|R_{d^o_k}|^{(1-d^o_k)}}.
\]
We then compute the overall odds of guessing correctly on which branch $M^{obs}$ belongs along the entire coding sequence $<d^o_1, d^o_2,...d^o_{K_o}>$ as
\[
PO(M^{obs})=\prod^{K_o}_{k=1}PO[d^o_k|M^{obs}].
\]
An example is illustrated in Figure \ref{071220_6}.

Likewise we compute an ensemble of odds $\{PO(M_b)\}^B_{b=1}$. Then we compute the p-value of observing an odds like $PO(M^{obs})$ as the proportion of $PO(M_b)$ being less than $PO(M^{obs})$:
\[
p(M^{obs})=\frac{\sum^B_{b=1}1_{[PO(M_b)< PO(M^{obs})]}}{B}.
\]
Upon the HC tree shown in Figure \ref{071120_(15-16)}(b), we have $PO(M^{obs})=0.023$ and p-value is $p(M^{obs})=0.002$. Hence, it turns out that $M^{obs}$ is significantly extreme in the HC tree. Based on the visible patterns observed in the heatmap, we can conclude that the 25 [Highly-dense] rectangles are not uniformly distributed.

\begin{figure}[ht!]
\centering
\begin{subfigure}[t]{.4\textwidth}
\centering
\includegraphics[width=0.8\linewidth]{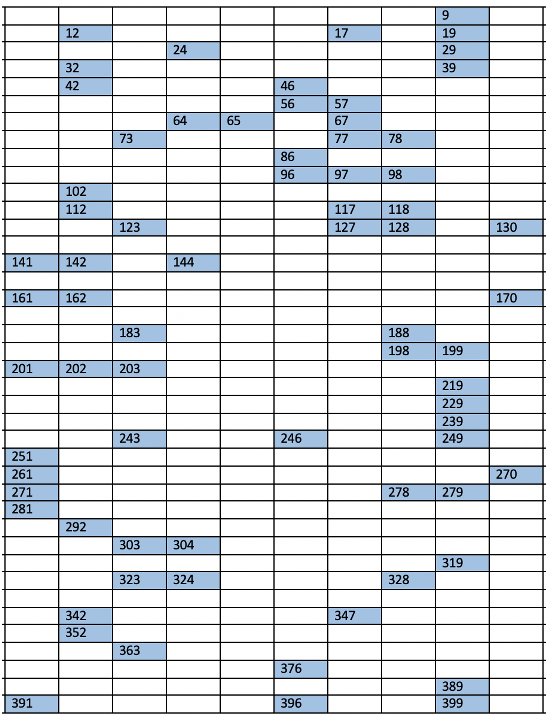}
\caption{}
\end{subfigure}
\hfill
\begin{subfigure}[t]{.4\textwidth}
\centering
\includegraphics[width=1\linewidth]{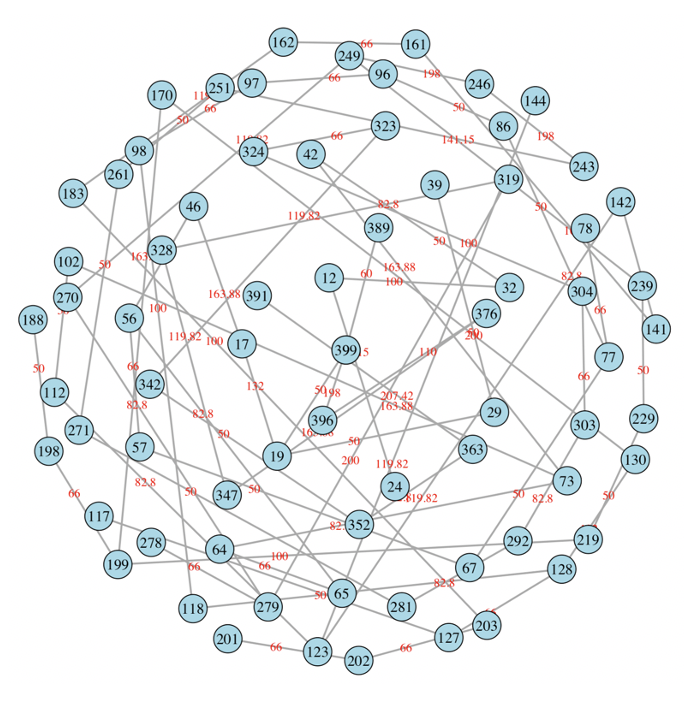}
\caption{}
\end{subfigure}
\captionsetup{justification=centering}
\caption{Enlarge the spatial scale: (a) focusing on 72 rectangles with 1 large dot or $\geq$ 1 medium dot; (b) the corresponding MST}
\label{071120_(17-18)}
\end{figure}

\begin{figure}[ht!]
\centering
\begin{subfigure}[t]{.45\textwidth}
\centering
\includegraphics[width=1\linewidth]{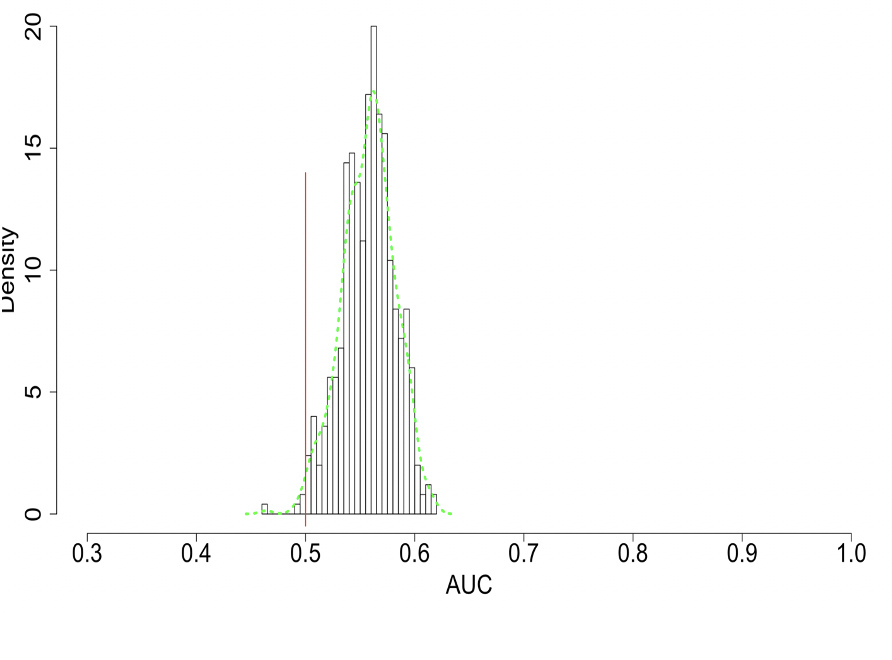}
\caption{}
\end{subfigure}
\hfill
\begin{subfigure}[t]{.45\textwidth}
\centering
\includegraphics[width=1\linewidth]{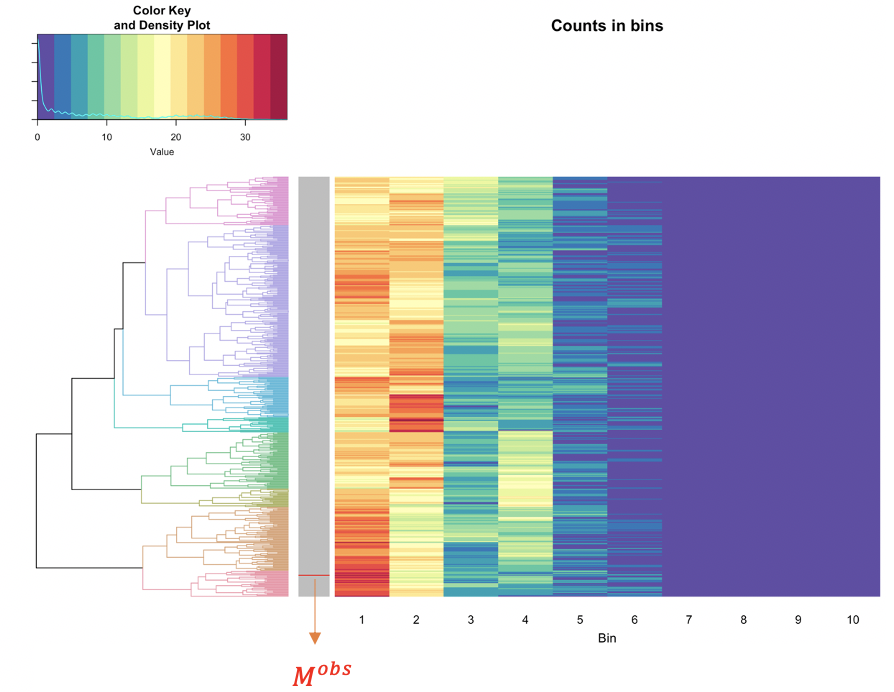}
\caption{}
\end{subfigure}
\captionsetup{justification=centering}
\caption{(a): ROC curve analysis results; (b): HC algorithm with hearmap, product of odds ($PO=1.108$) and p-value $=0.056$.}
\label{071120_(19-20)}
\end{figure}

\clearpage
\paragraph{Enlarge the spatial scale:}
In the previous section, we have built a MST for the Highly-dense rectangles' 2D-distribution, as the 3rd of the flow chart shown in Figure \ref{071120_(8-12)}. We further work on 2D distribution of the 4th one including those Dense rectangles, as shown in Figure \ref{071120_(17-18)}(a). By combining the [R1] and [R2] scales of rectangles, we expect to see the distribution of purple-dots with an expanded perspective. Its MST is computed and reported in Figure \ref{071120_(17-18)}(b). Likewise, we perform the two versions of testing on spatial uniformness and report the results in Figure \ref{071120_(19-20)}, respectively.
Though having less significance in terms of p-values, the results via ROC curve analysis on the left panel and Machine Learning method on the right panel all still indicate that the rectangles' 2D distribution is not fully in accord with spatial uniformness.

Nonetheless, this trend of getting less and less significant in against spatial uniformness is expected when we further expand by including rectangles of [R3] Sparse scale. This trend tells us that the spraying mechanism needs further fine tuning in order to achieve spatial uniformness. Especially, large purple nodes in [R1] Highly-dense scale should be significantly reduced.

\section{Analysis of 4 other images}
We apply computational algorithms developed and illustrated through image no.1 to the rest of four images. Heterogeneous shading conditions can be evidently seen across these four images. Overall our exhaustive color identifications are satisfactory and the testings of spatial uniformness are indeed much more effective than the one based on ROC analysis.

\subsection{Image no.2}
\begin{figure}[ht!]
\centering
\begin{subfigure}[t]{.3\textwidth}
\centering
\includegraphics[height=7cm, width=5cm]{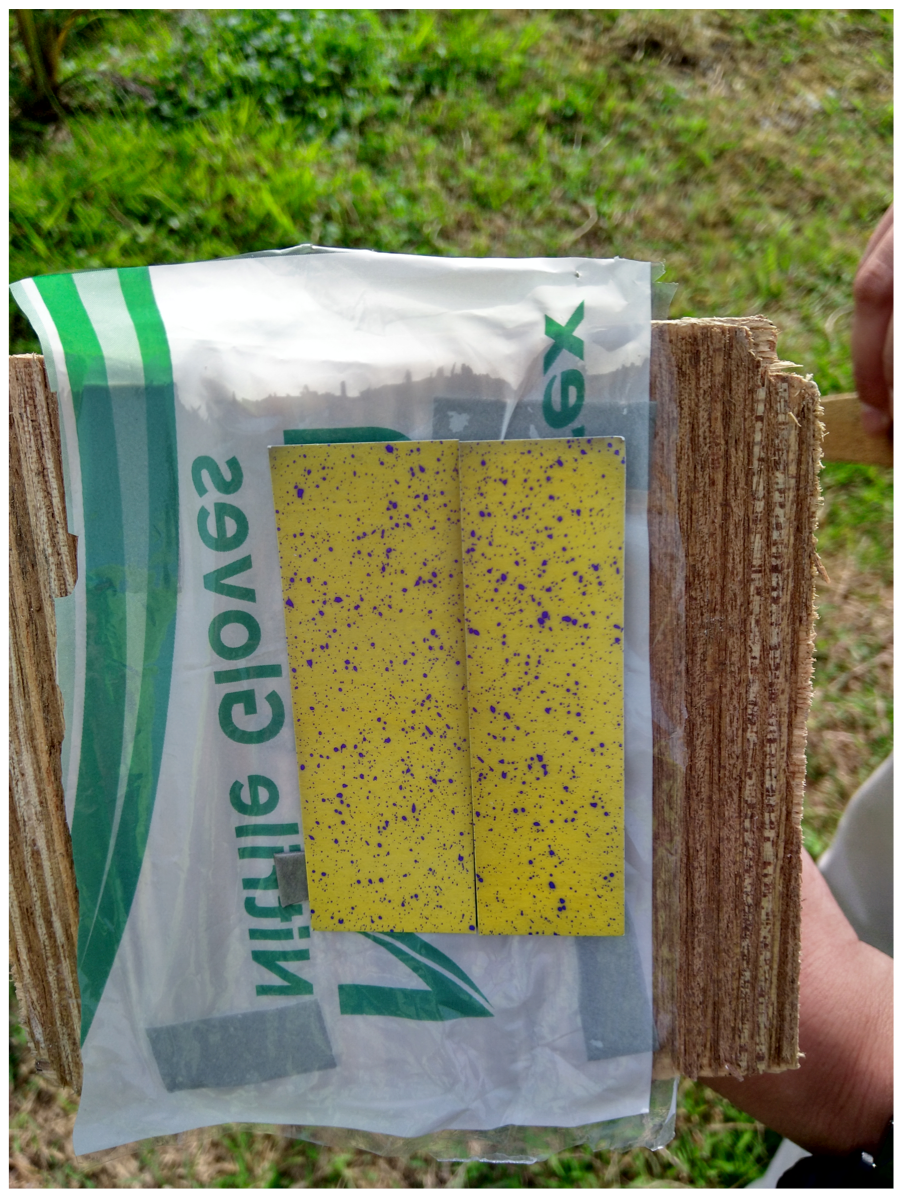}
\caption{}
\end{subfigure}
\hfill
\begin{subfigure}[t]{.3\textwidth}
\centering
\includegraphics[height=7cm, width=5cm]{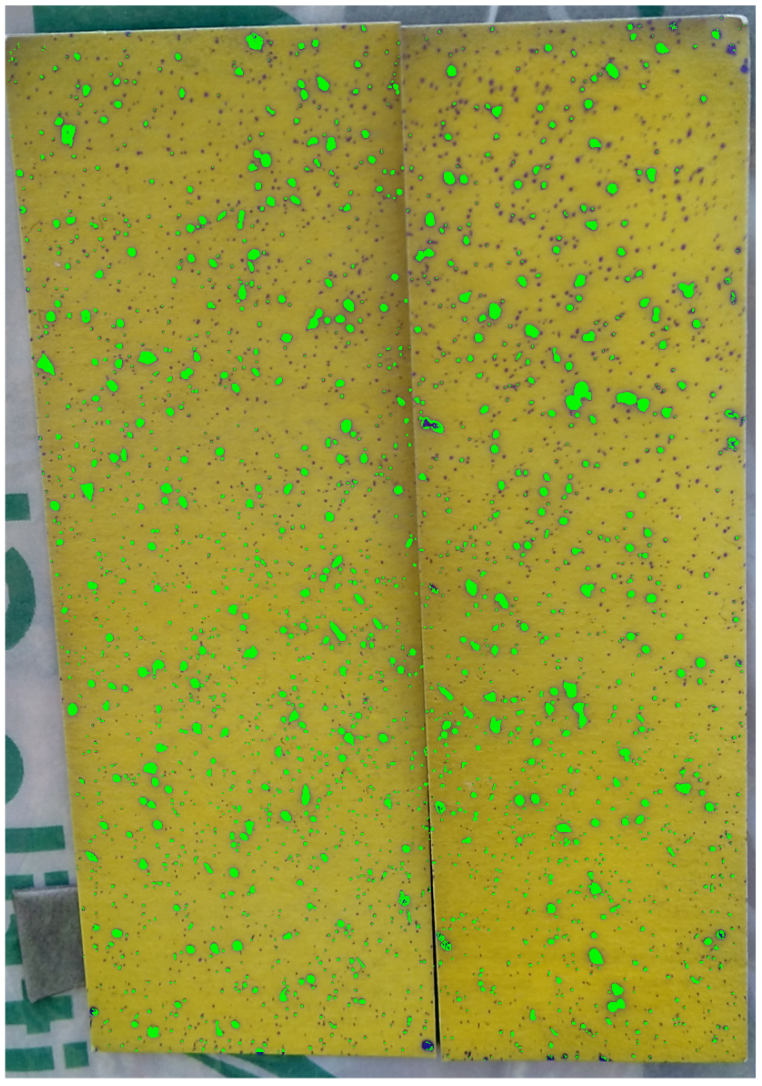}
\caption{}
\end{subfigure}
\hfill
\begin{subfigure}[t]{.3\textwidth}
\centering
\includegraphics[height=7cm, width=5cm]{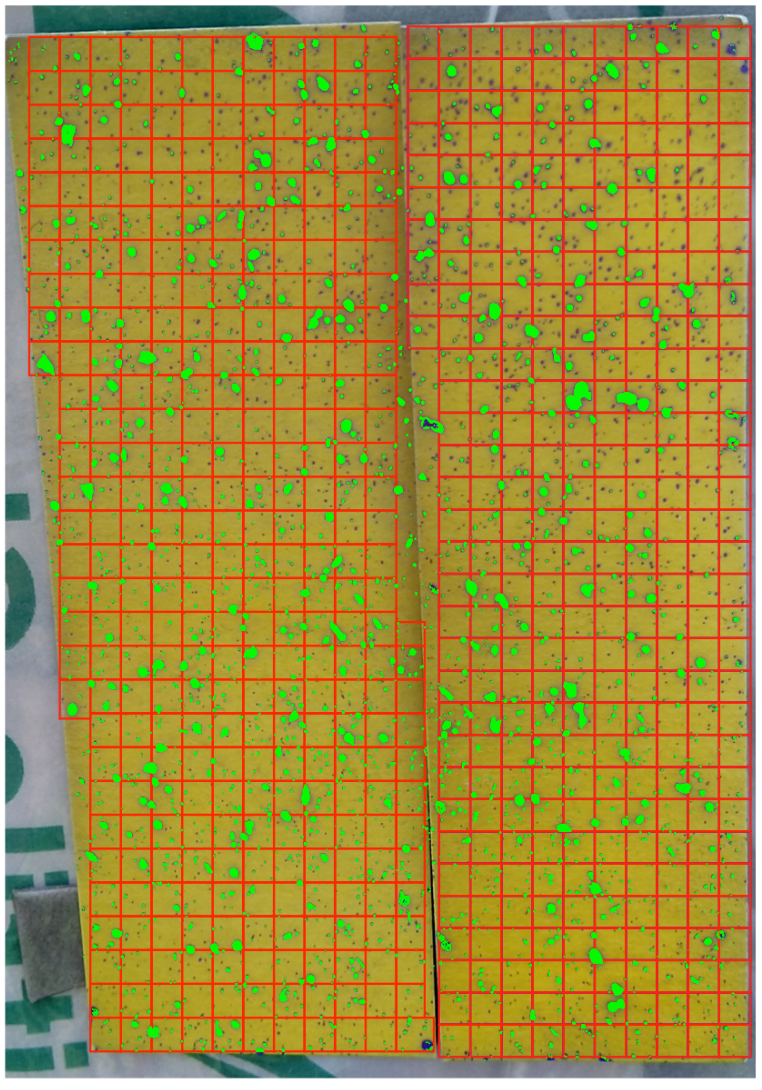}
\caption{}
\end{subfigure}
\captionsetup{justification=centering}
\caption{Image no.2: (a) two test papers in the original data; (b) recovering by RGB file; (c) dividing the paper into rectangles for 2D spatial uniformness testing.}
\label{072420_i2a}
\end{figure}

The image no.2 consists of two ``pages'' of test papers, as shown in Figure \ref{072420_i2a}(a). The upper part of these two coupled test papers is under shading. Consequently, the color-dot identification based only RGB has missed quite a few small purple dots, as shown in Figure \ref{072420_i2a}(b). Many of these small dots were also not been picked up via HSV data format based on $1\times 1 \times 1$ fine scale cubes.

\begin{figure}[ht!]
\centering
\begin{subfigure}[t]{.45\textwidth}
\centering
\includegraphics[width=1\linewidth]{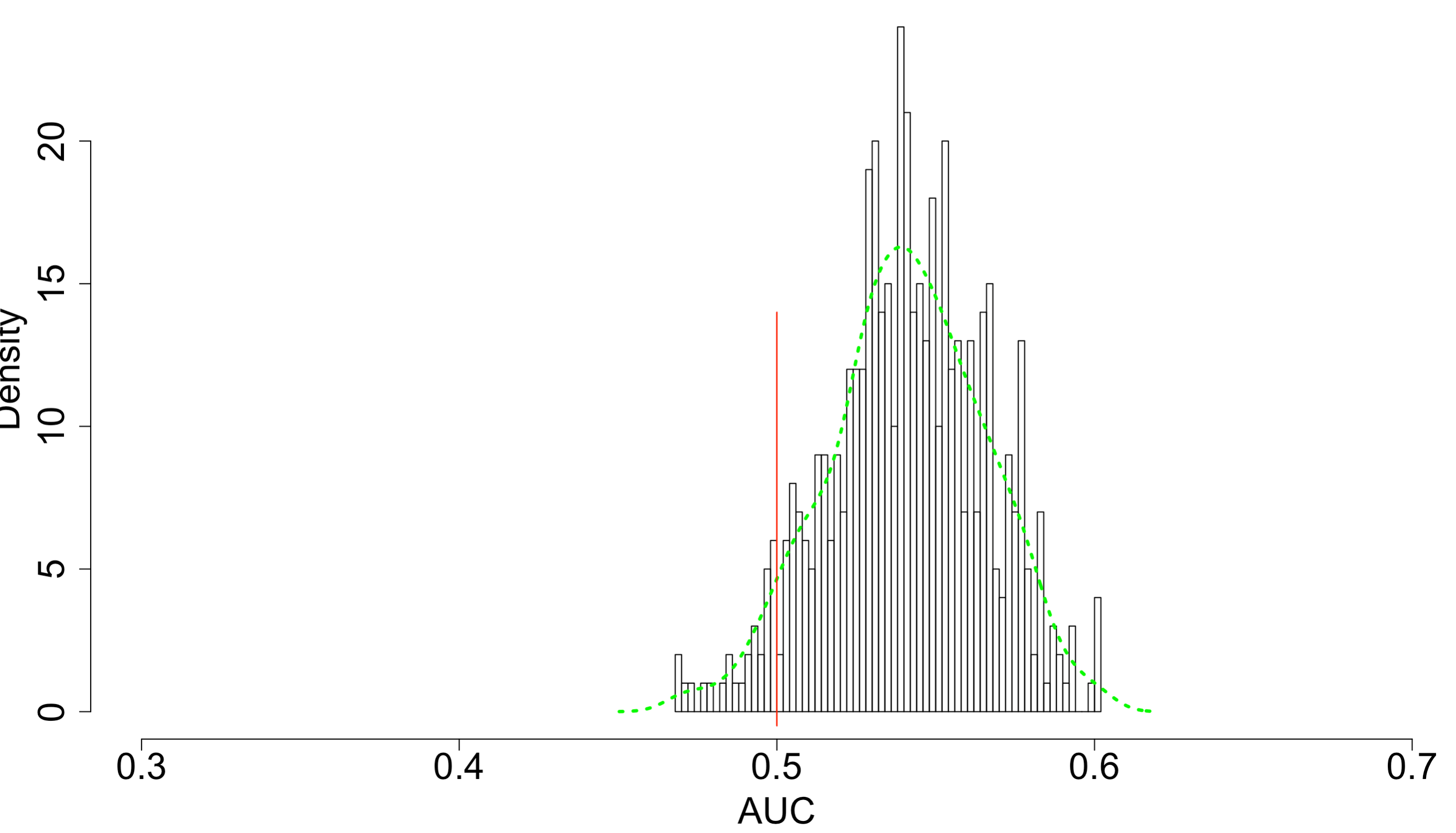}
\caption{}
\end{subfigure}
\hfill
\begin{subfigure}[t]{.45\textwidth}
\centering
\includegraphics[width=1\linewidth]{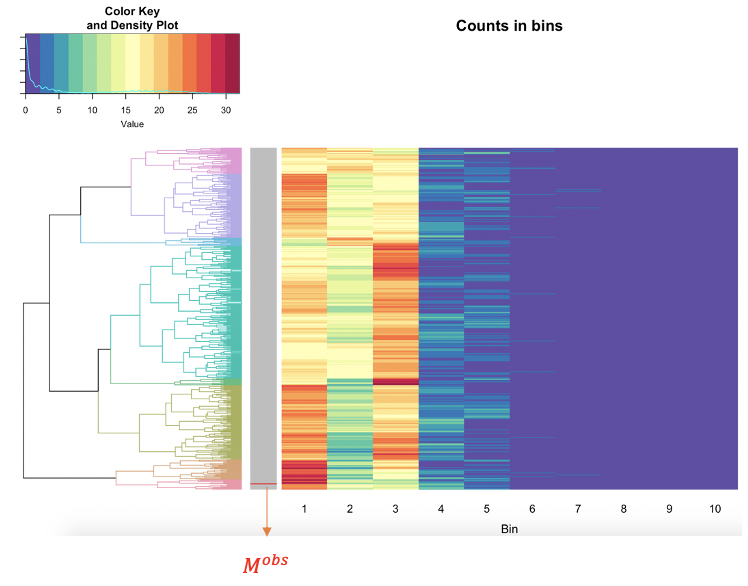}
\caption{}
\end{subfigure}
\captionsetup{justification=centering}
\caption{Image no.2: Spatial uniformness testing on the LEFT test paper: studying 62 rectangles including 55 rectangles with $\geq$ 2 large dots and 7 rectangles with 1 large dot and $\geq 2$ medium dots; (a): ROC curve anlysis results; (b): HC algorithm with heatmap, product of odds ($PO=0.016$) and p-value $p(M^{obs})=0$.}
\label{072420_i2b}
\end{figure}

\begin{figure}[ht!]
\centering
\begin{subfigure}[t]{.45\textwidth}
\centering
\includegraphics[width=1\linewidth]{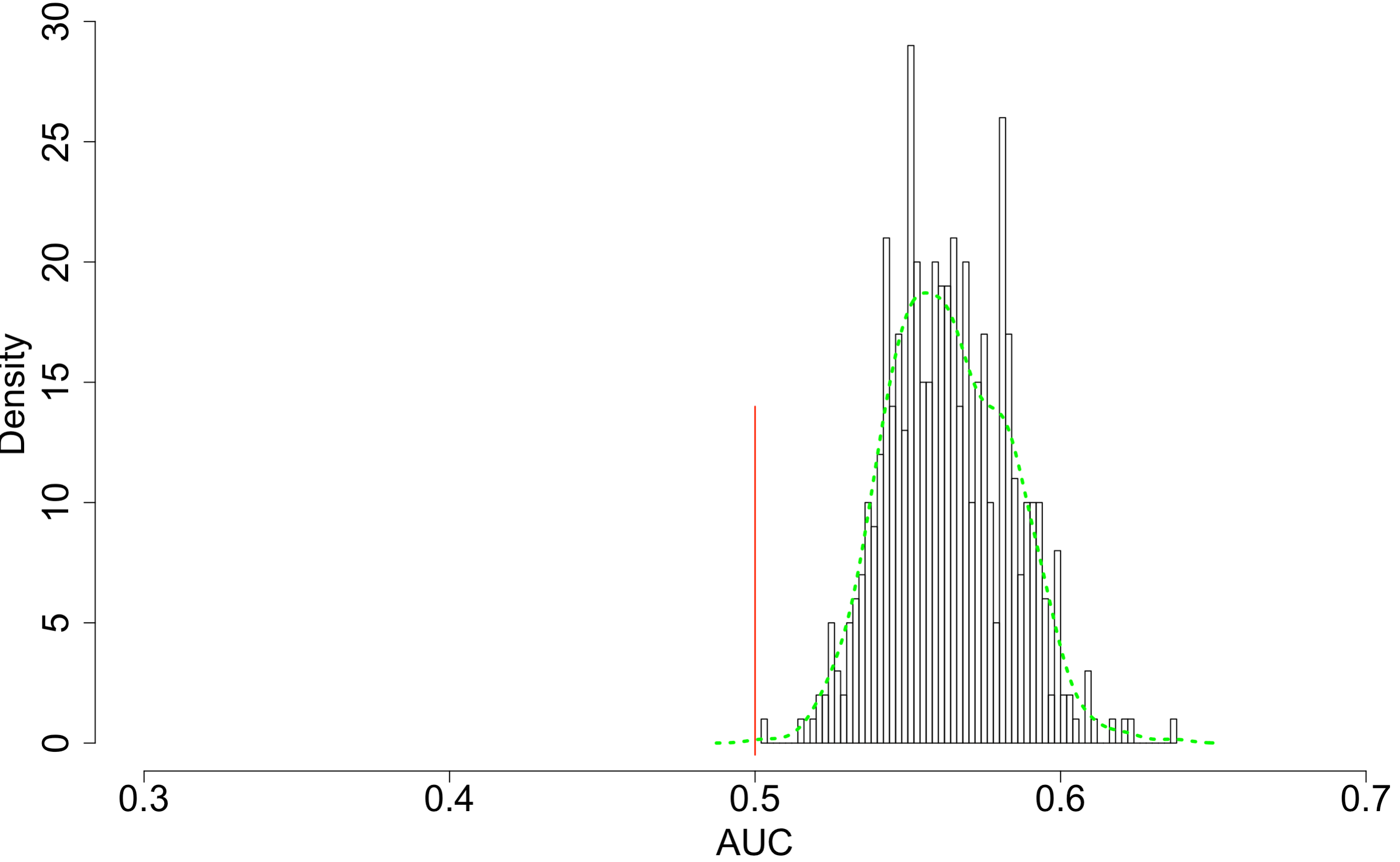}
\caption{}
\end{subfigure}
\hfill
\begin{subfigure}[t]{.45\textwidth}
\centering
\includegraphics[width=1\linewidth]{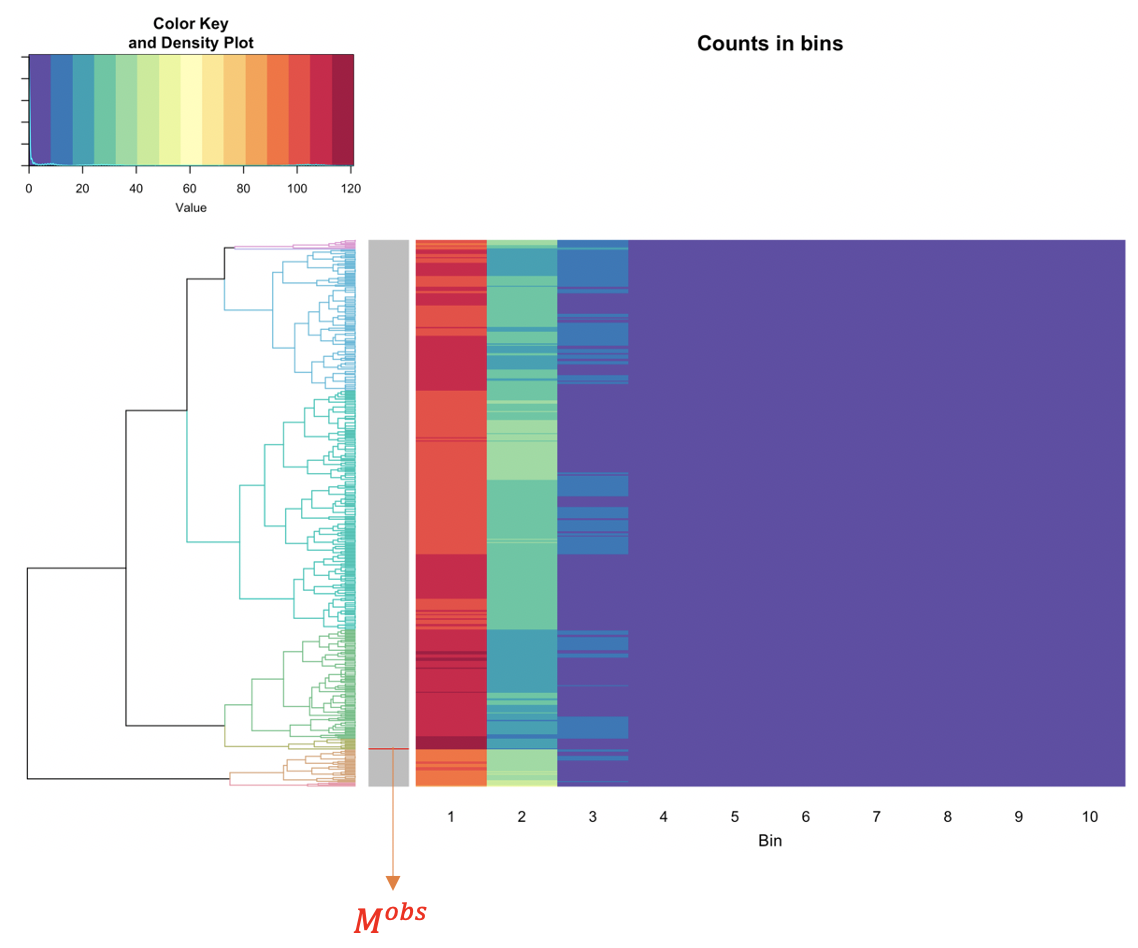}
\caption{}
\end{subfigure}
\captionsetup{justification=centering}
\caption{Image no.2: Spatial uniformness testing on the RIGHT test paper: studying 142 rectangles with $\geq$ 1 large dots; (a): ROC curve anlysis results; (b): HC algorithm with heatmap, product of odds ($PO=0.047$) and p-value $p(M^{obs})=0.004$.}
\label{072420_i2c}
\end{figure}

We separately report results of spatial uniformness on the two test papers by focusing only dense squares as shown in Figure \ref{072420_i2a}(c). Two separate results are reported: Figure \ref{072420_i2b} for the Left and Figure \ref{072420_i2c} for the Right, respectively. Based on both figures, we see that there exits a small discrepancy in p-values between the result based on ROC in Figure \ref{072420_i2b}(a) against results based on HC-tree and heatmap in Figure \ref{072420_i2b}(b), and result based on ROC in Figure \ref{072420_i2c}(a) against results based on HC-tree and heatmap in Figure \ref{072420_i2c}(b). However, such small discrepancies don't seem to cause incoherent conclusions.

\subsection{Image no.3}
\begin{figure}[ht!]
\centering
\begin{subfigure}[t]{.3\textwidth}
\centering
\includegraphics[height=7cm, width=5cm]{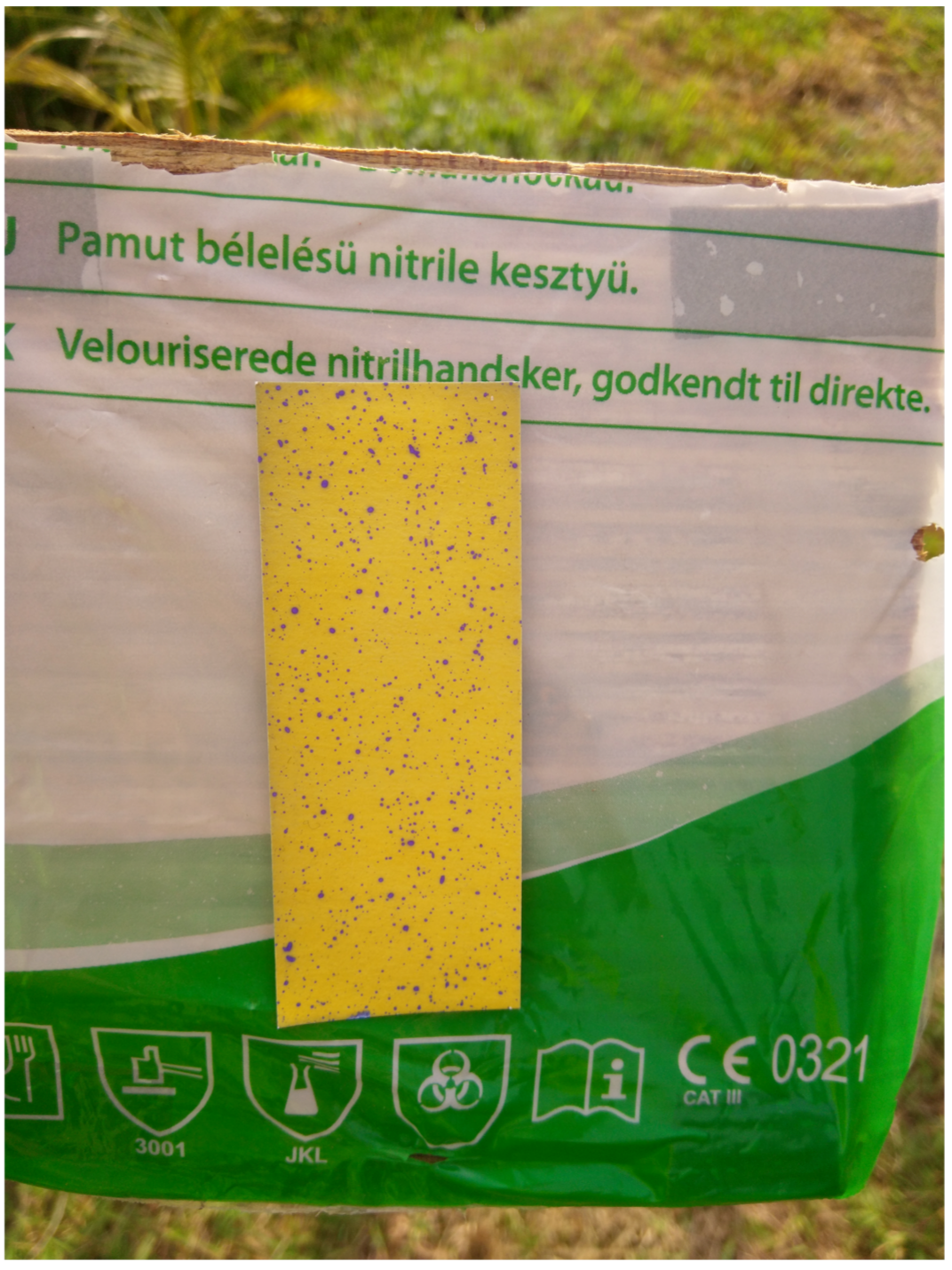}
\caption{}
\end{subfigure}
\hfill
\begin{subfigure}[t]{.3\textwidth}
\centering
\includegraphics[height=7cm, width=3.5cm]{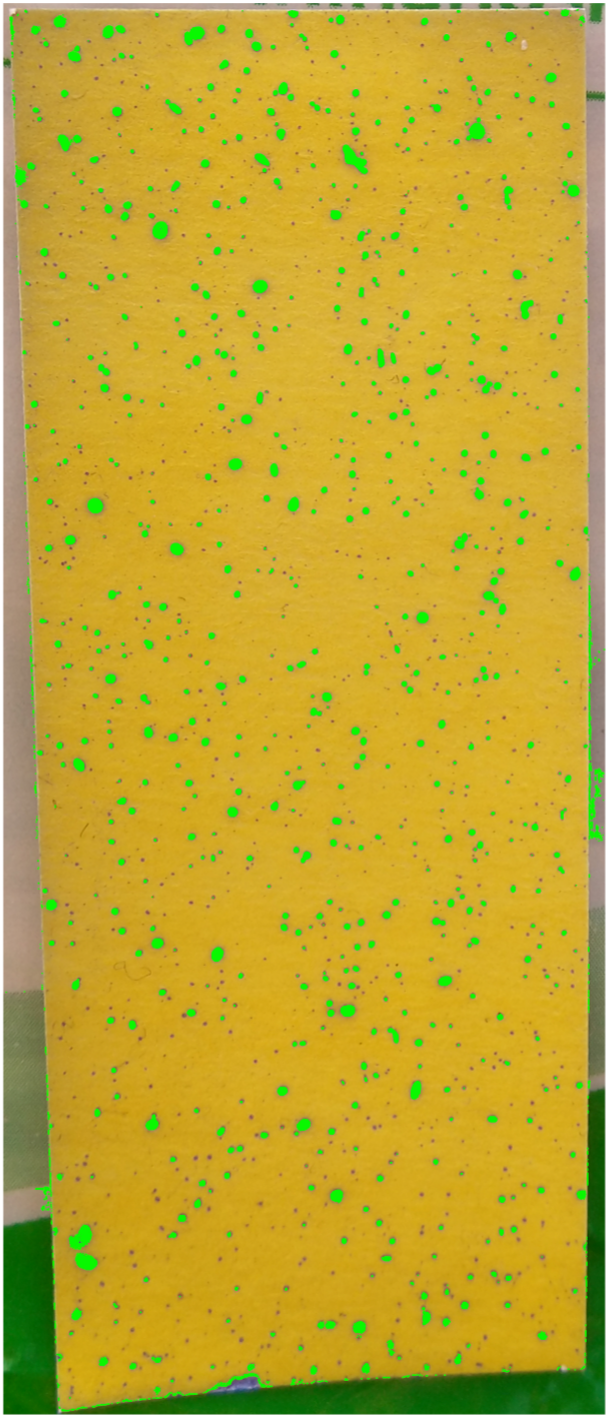}
\caption{}
\end{subfigure}
\hfill
\begin{subfigure}[t]{.3\textwidth}
\centering
\includegraphics[height=7cm, width=3.5cm]{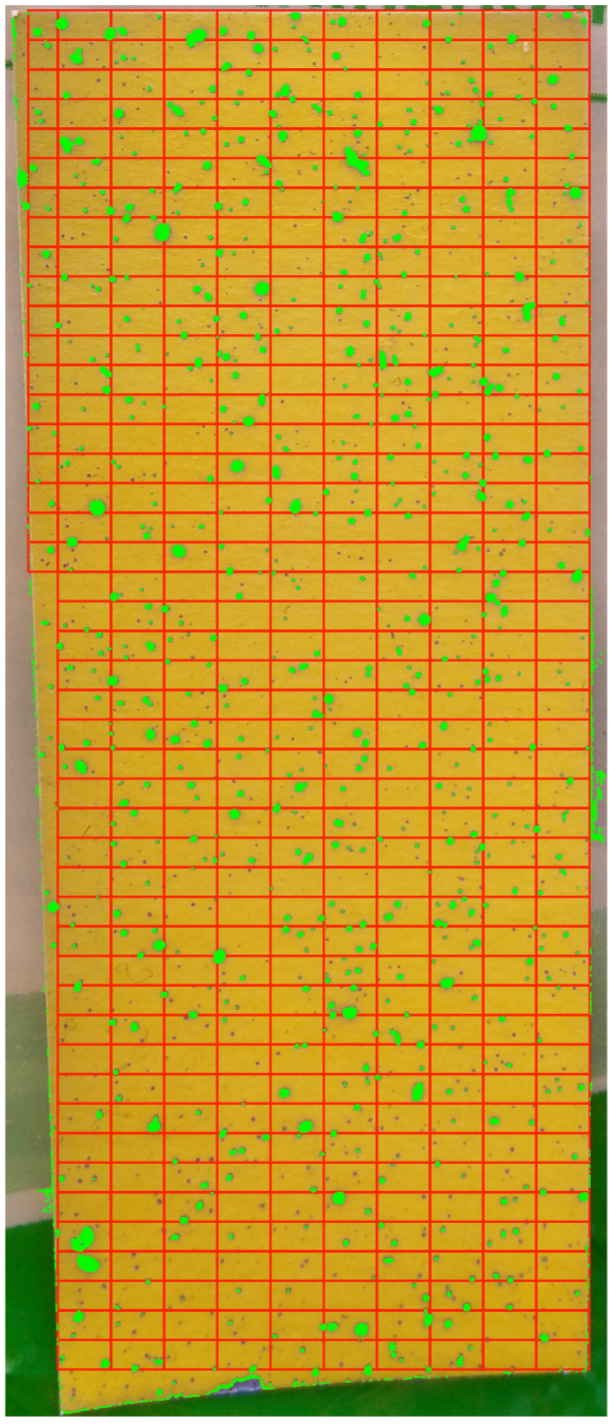}
\caption{}
\end{subfigure}
\captionsetup{justification=centering}
\caption{Image no.3:(a) one test paper in the original data; (b) recovering by 6D [RGB+HSV] file ($n=10$); (c) dividing the paper into 479 rectangles for 2D spatial uniformness testing.}
\label{072520_i3a}
\end{figure}

\begin{figure}[ht!]
\centering
\begin{subfigure}[t]{.45\textwidth}
\centering
\includegraphics[width=1\linewidth]{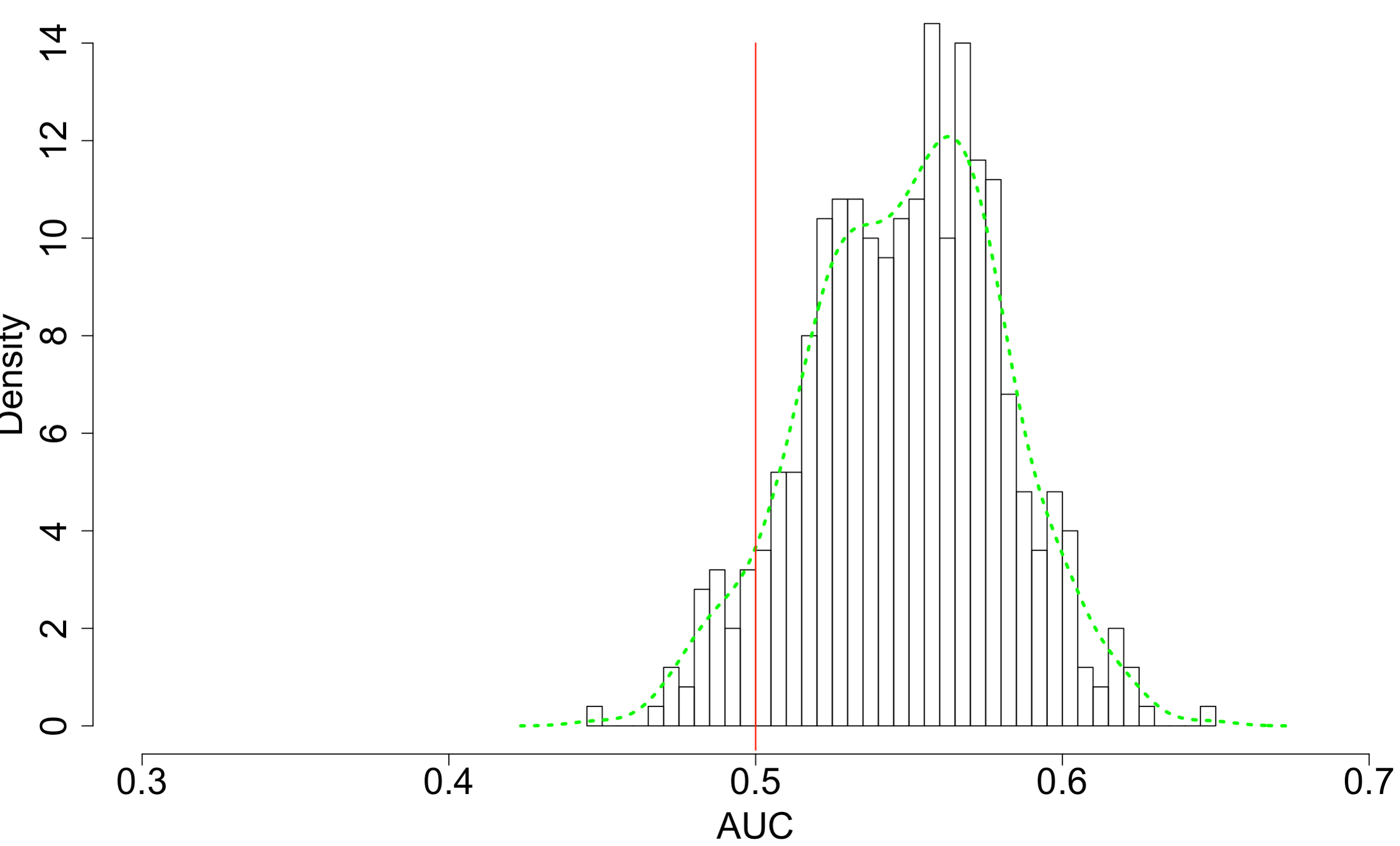}
\caption{}
\end{subfigure}
\hfill
\begin{subfigure}[t]{.45\textwidth}
\centering
\includegraphics[width=1\linewidth]{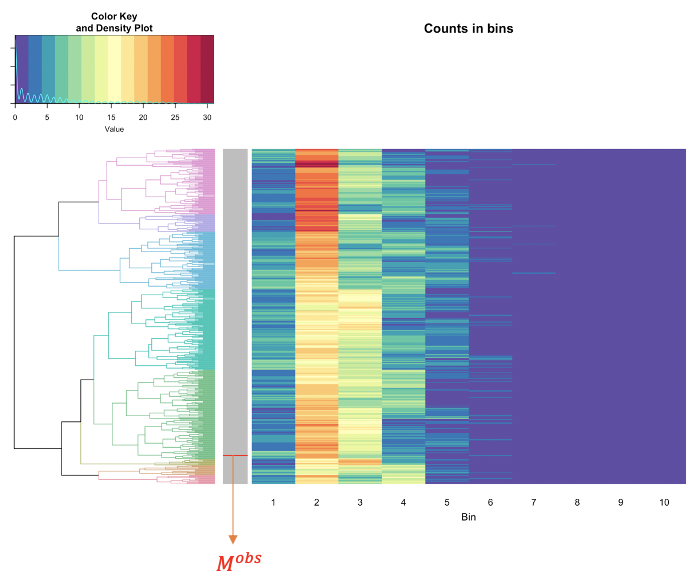}
\caption{}
\end{subfigure}
\captionsetup{justification=centering}
\caption{Image no.3: Spatial uniformness testing based on 49 rectangles including 43 rectangles with $\geq$ 1 large dots and 6 rectangles with $\geq 4$ medium dots;(a): ROC curve anlysis results; (b): HC algorithm with heatmap, product of odds ($PO=1164.056$) and p-value $p(M^{obs})=0.686$.}
\label{072520_i3b}
\end{figure}

The test paper in image no.3 seems curved a bit, as shown in Figure \ref{072520_i3a}(a). This curved shape likely created shads around upper left and lower right corners of the test paper. The coupled results from RGB and HSV seem to achieve a big degree of exhaustive identification except dots locating the two corners, as shown in Figure \ref{072520_i3a}(b). For the spatial uniformness test, the result based on ROC analysis, as shown in Figure \ref{072520_i3b}(a), seems to point to a direction being slightly different from the one based on HC-tree and heatmap, as shown in Figure \ref{072520_i3b}(b). Given the observed row being well mixed with simulated one within a big branch, we are more confident on the p-value result based on HC-tree and heatmap Figure \ref{072520_i3b}(b)

\subsection{Image no.4}
\begin{figure}[ht!]
\centering
\begin{subfigure}[t]{.3\textwidth}
\centering
\includegraphics[height=7cm, width=5cm]{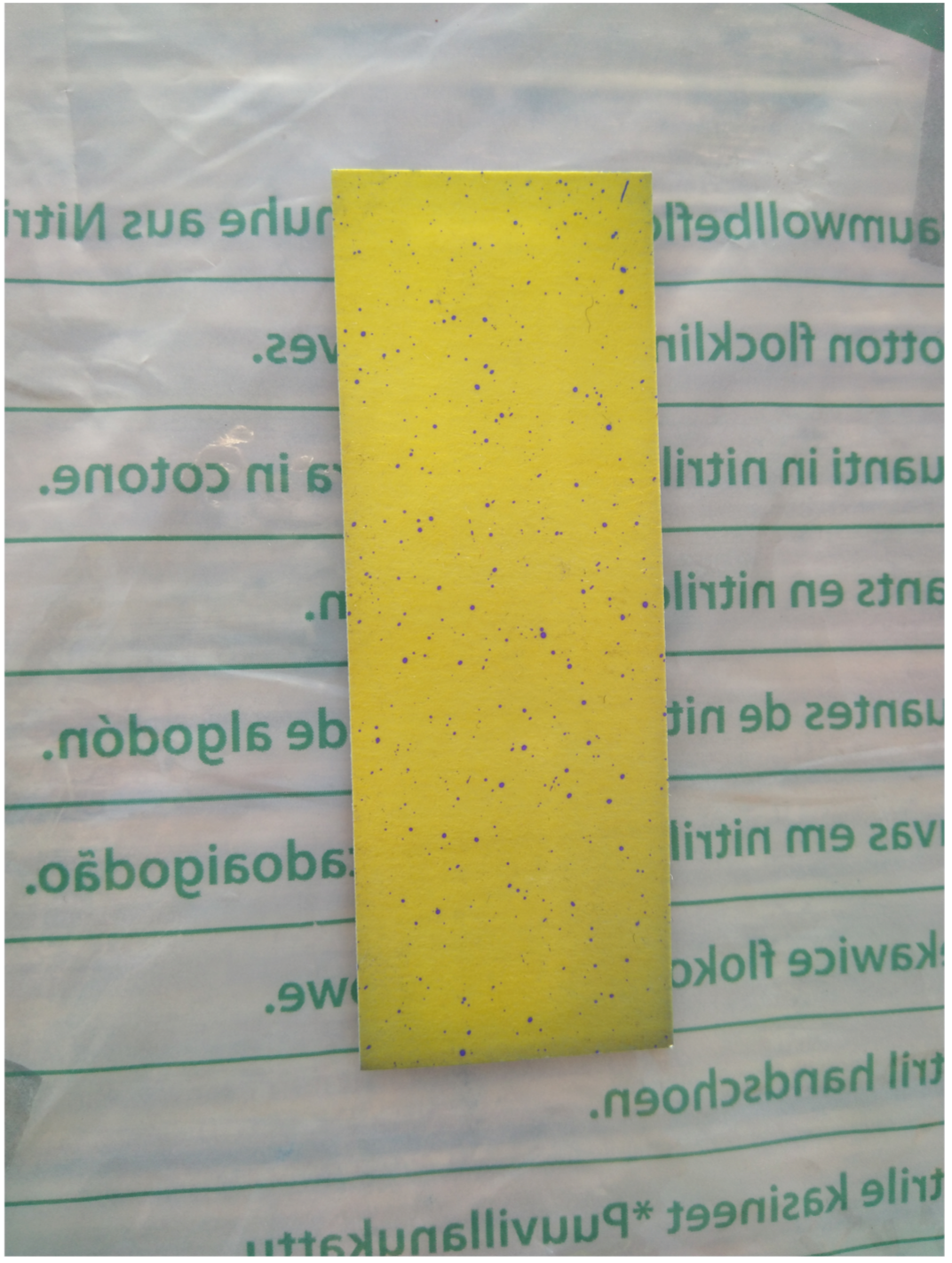}
\caption{}
\end{subfigure}
\hfill
\begin{subfigure}[t]{.3\textwidth}
\centering
\includegraphics[height=7cm, width=3.5cm]{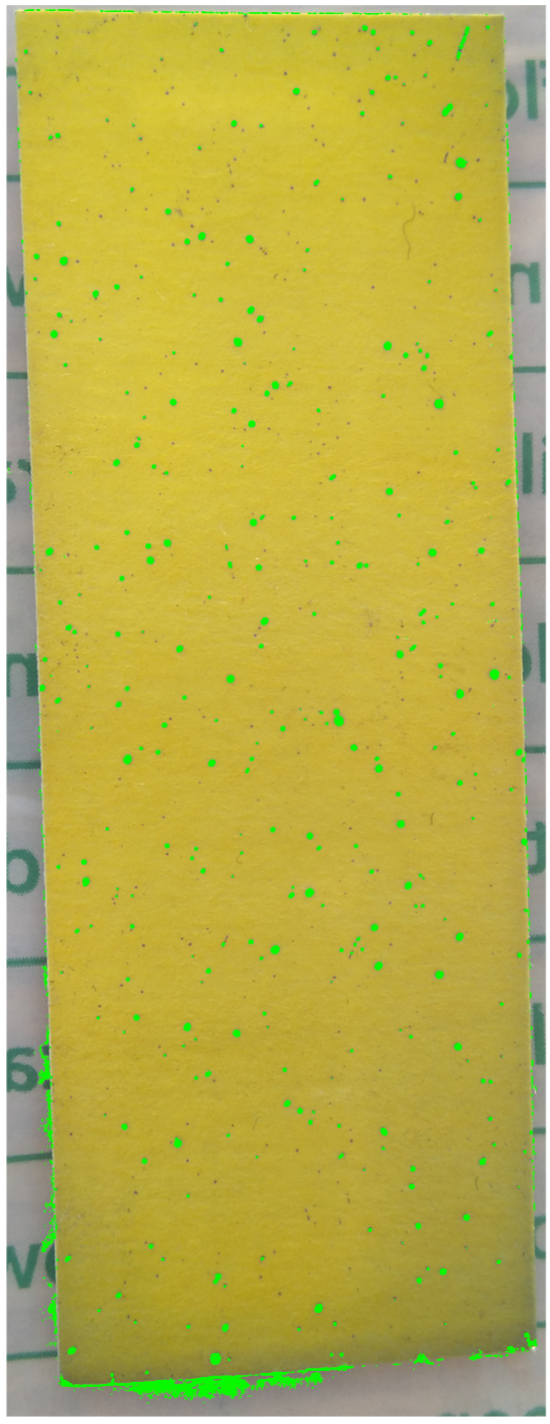}
\caption{}
\end{subfigure}
\hfill
\begin{subfigure}[t]{.3\textwidth}
\centering
\includegraphics[height=7cm, width=3.5cm]{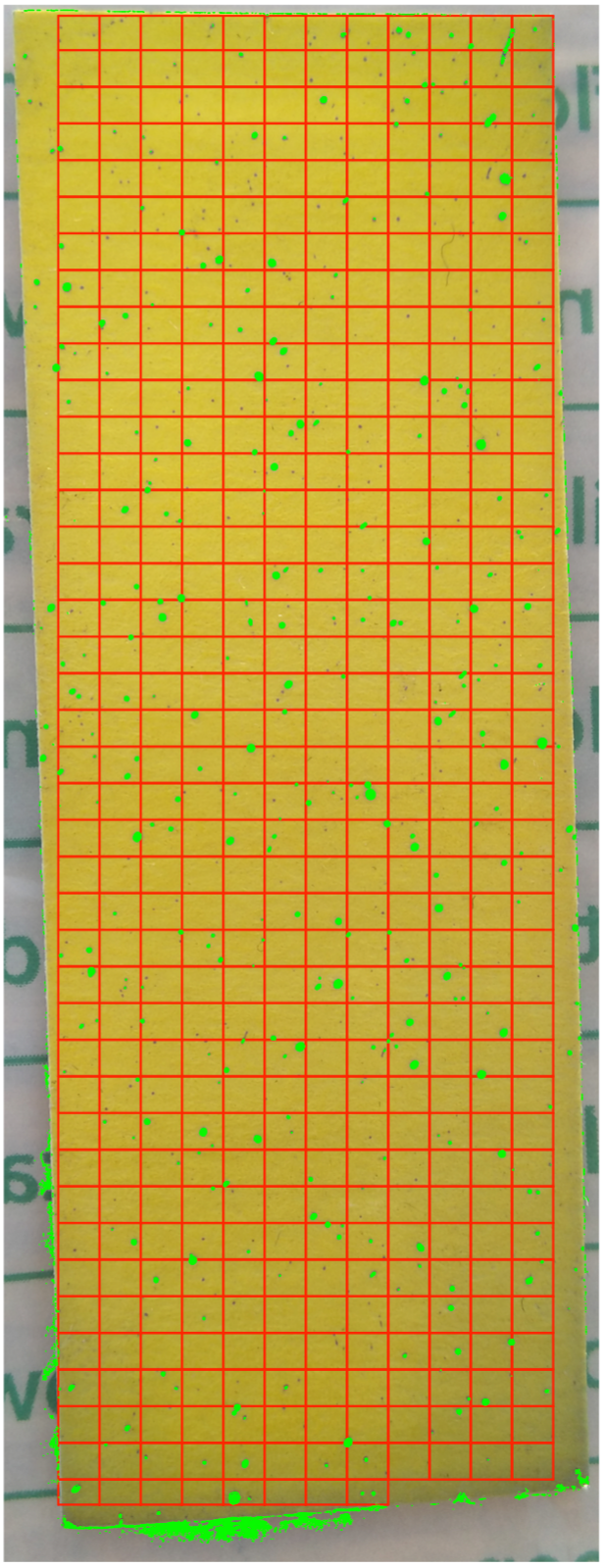}
\caption{}
\end{subfigure}
\captionsetup{justification=centering}
\caption{Image no.4: (a) one test paper in the original data; (b) recovering by HSV file ($n=10$); (c) dividing the paper into 488 rectangles for 2D spatial uniformness testing.}
\label{072620_i4a}
\end{figure}

\begin{figure}[ht!]
\centering
\begin{subfigure}[t]{.45\textwidth}
\centering
\includegraphics[width=1\linewidth]{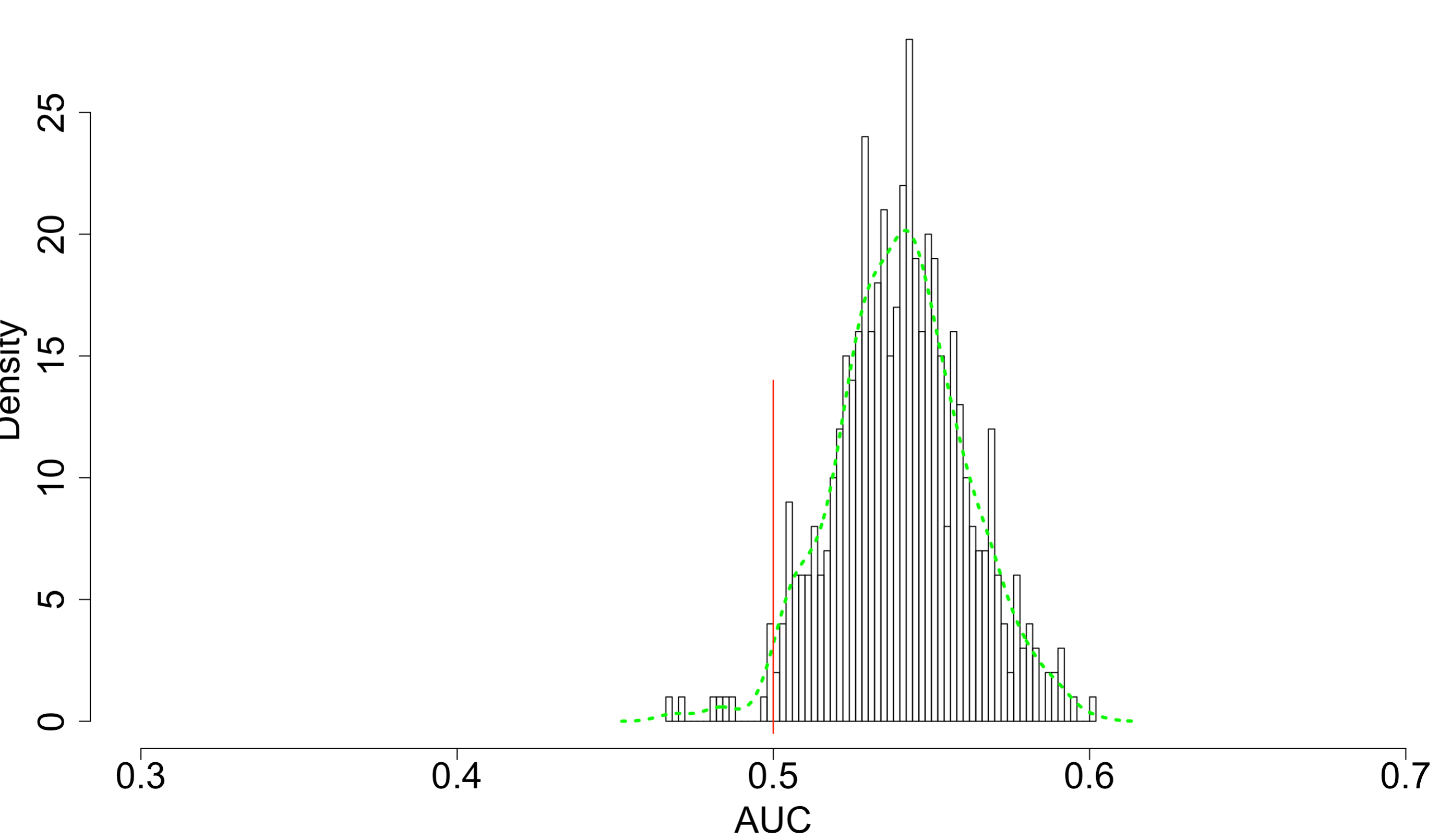}
\caption{}
\end{subfigure}
\hfill
\begin{subfigure}[t]{.45\textwidth}
\centering
\includegraphics[width=1\linewidth]{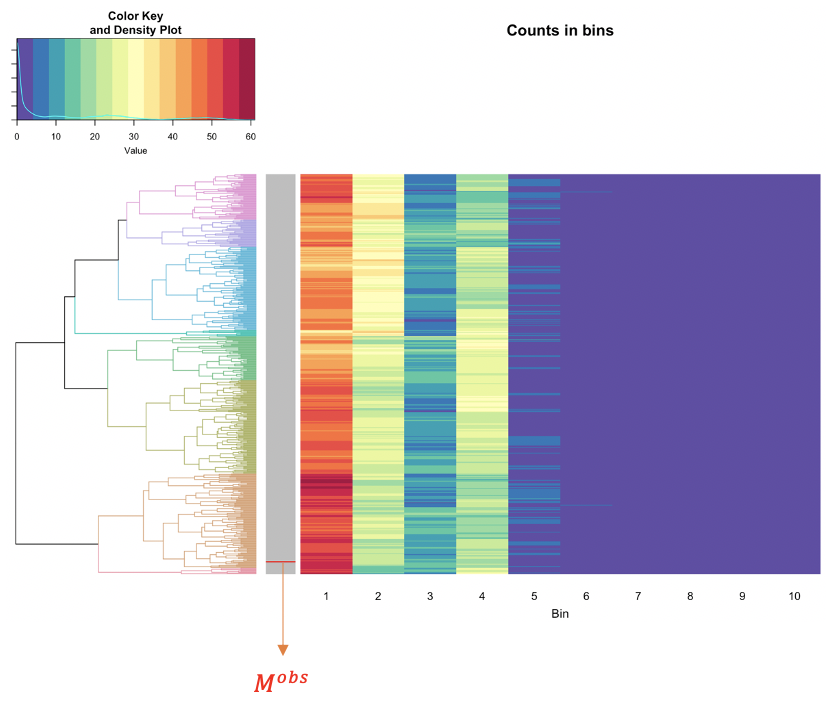}
\caption{}
\end{subfigure}
\captionsetup{justification=centering}
\caption{Image no.4: Spatial uniformness testing based on 112 rectangles with $\geq$ 1 large dots;(a): ROC curve anlysis results; (b): HC algorithm with heatmap, product of odds ($PO=193.157$) and p-value $p(M^{obs})=0.642$.}
\label{072520_i4b}
\end{figure}

The image no.4 have obvious shades at the upper and lower boundaries of the test paper, as shown in Figure \ref{072620_i4a}(a). We report the color-dot identification result based on HSV cubes of $n=10$ scale, as shown in Figure \ref{072620_i4a}(b). For spatial uniformness testing, a big gap is seen between the result based on ROC analysis and the one based on HC-tree and Heatmap, as shown in Figure \ref{072520_i4b}(a) and (b), respectively. Again the latter result seems more reliable.

\subsection{Image no.5}
\begin{figure}[ht!]
\centering
\begin{subfigure}[t]{.3\textwidth}
\centering
\includegraphics[height=7cm, width=5cm]{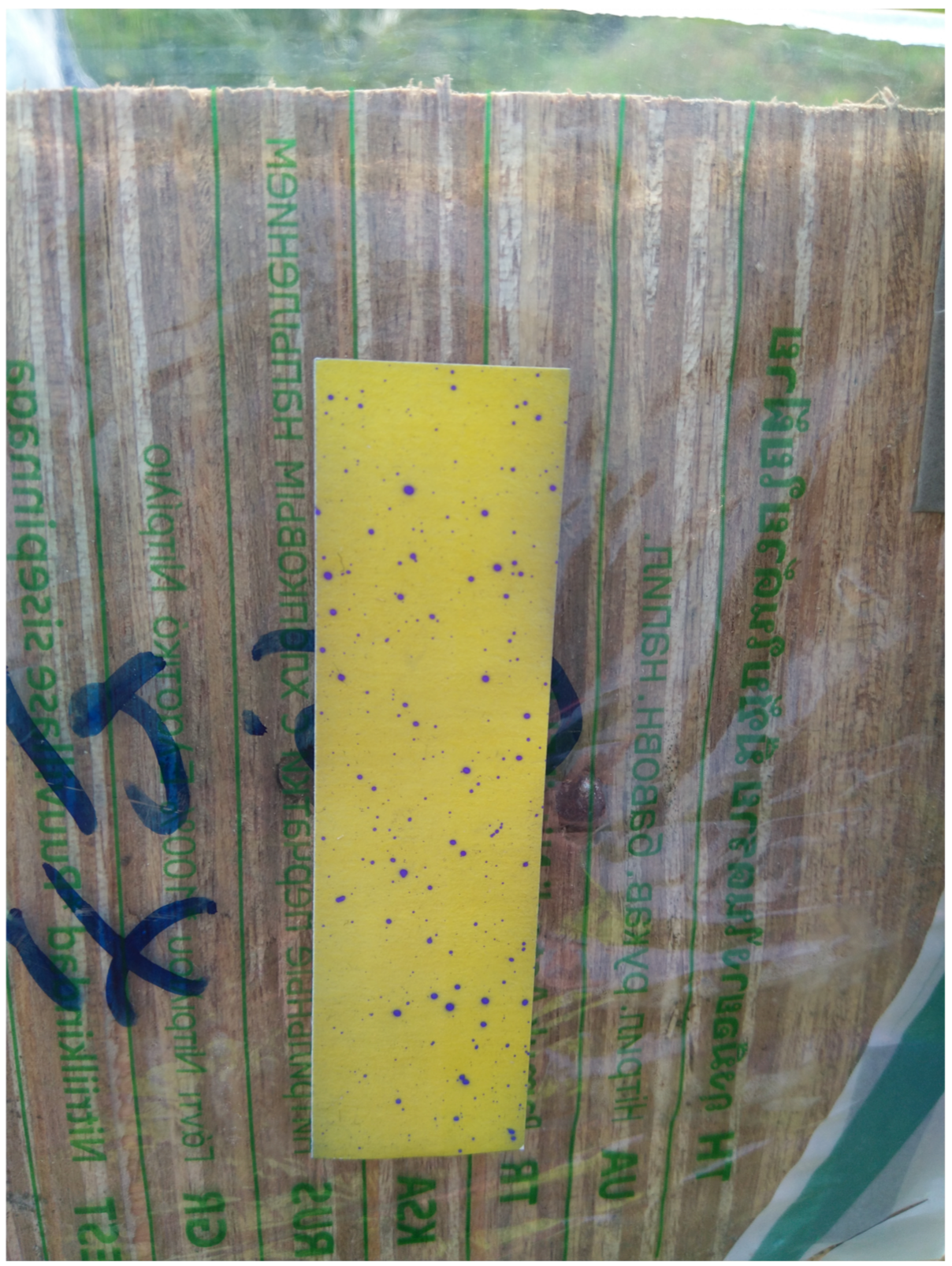}
\caption{}
\end{subfigure}
\hfill
\begin{subfigure}[t]{.3\textwidth}
\centering
\includegraphics[height=7cm, width=3.5cm]{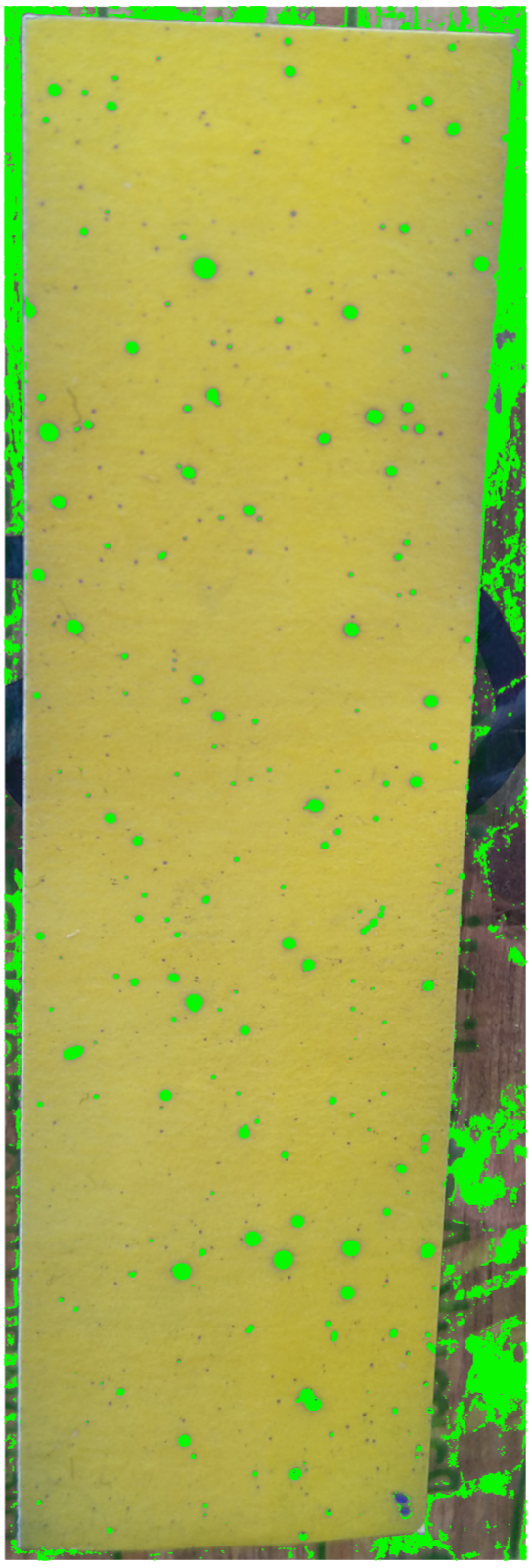}
\caption{}
\end{subfigure}
\hfill
\begin{subfigure}[t]{.3\textwidth}
\centering
\includegraphics[height=7cm, width=3.5cm]{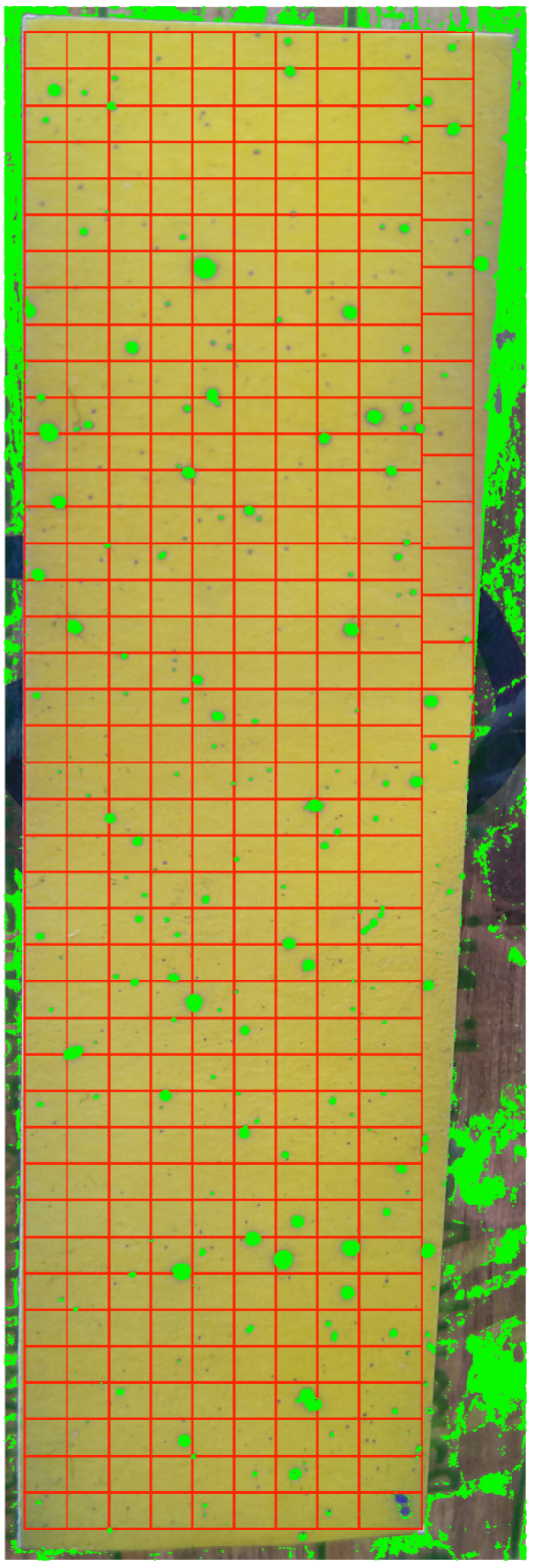}
\caption{}
\end{subfigure}
\captionsetup{justification=centering}
\caption{Image no.5: (a) one test paper in the original data; (b) recovering by 6D [RGB+HSV] file ($n=10$); (c) dividing the paper into 384 rectangles for 2D spatial uniformness testing.}
\label{072620_i5a}
\end{figure}

\begin{figure}[ht!]
\centering
\begin{subfigure}[t]{.45\textwidth}
\centering
\includegraphics[width=1\linewidth]{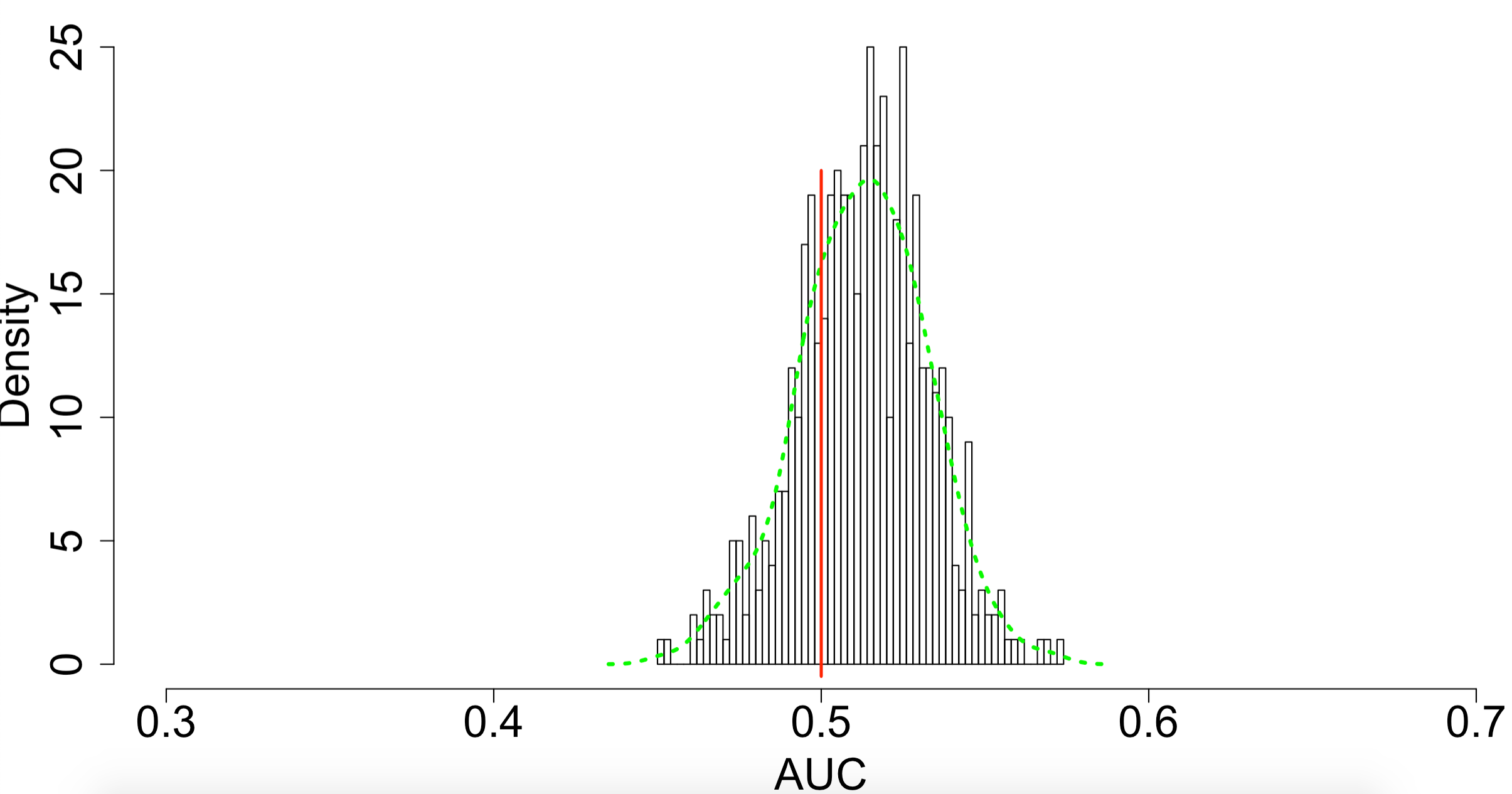}
\caption{}
\end{subfigure}
\hfill
\begin{subfigure}[t]{.45\textwidth}
\centering
\includegraphics[width=1\linewidth]{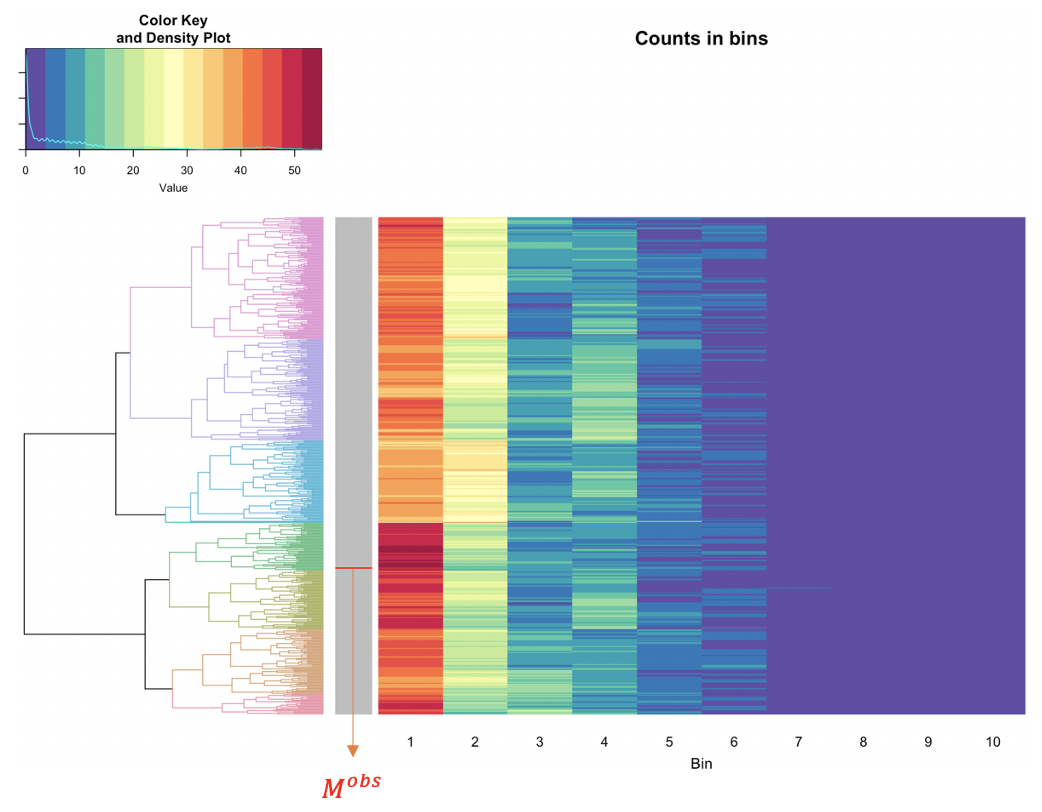}
\caption{}
\end{subfigure}
\captionsetup{justification=centering}
\caption{Image no.5: Spatial uniformness testing based on 97 rectangles with $\geq$ 1 large dots;(a): ROC curve anlysis results; (b): HC algorithm with heatmap, product of odds ($PO=0.334$) and p-value $p(M^{obs})=0.056$.}
\label{072620_i5b}
\end{figure}

The image no.5 looks like being twisted a bit, particularly at the lower right corner, and the shades are visible all over, as shown in Figure \ref{072620_i5a}(a). The final color-dot identification is based on the coupled results of RGB and HSV, as shown in Figure \ref{072620_i5a}(b). The discrepancy between the two results of spatial uniformness testing is especially wide. But, based on the small size of the branch containing the observed row vector, we have more confidence on the one based on HC-tree and heatmap, as shown in Figure \ref{072620_i5b}(b), over the ROC one, as shown in Figure \ref{072620_i5b}(a).

\section{Conclusions}
The color-identifications and testing 2D spatial uniformness via MST for the five images are rather satisfactory. Basically, these results collectively strongly indicate that our data-driven computational approach for color-identifications are rather effective, and testing methodology for 2D spatial uniformness is novel and practical.

The underlying reason of the effectiveness of our color-identification approach is the low color-complexity. This interesting fact is that this simple concept is not well known in literature. In fact, our current color research has shown us that low color-complexity is seen in natural images as well as in images of famous paintings. That is, our data-driven color-identification is applicable in wide range of color images.

The MST structure and its distance distribution are new and essential summarizing pattern information of spatial data. They fit very well into this Data Sciences. The novelty of evaluating p-value via products of odds-ratios based on a tree structure, which is complex, can critically expand the applications of unsupervised machine learning methodologies to wider ranges of scientific fields, including the medical one.

Our way of dealing with shading in images is not sophisticated. We adopt the fact that RGB and HSV data formats could be differentially affected by shading. So, we propose to combine results from both data formats under distinct scales. From the results reported in Section 5, we see that it works with different degrees of successes. It might be possible to develop systematic approaches to remove, or at least lessen the shading effects. This is one of our undertaking research directions right now.



\bibliographystyle{unsrt}

\begin{thebibliography}{1}

    \bibitem{faical}
    Faical, B. S., Costa, F. G., Pessin, G., Ueyama, J., Freitas, H., Colombo, A., Fini, P. H., Villas, L. A., Osorio, F. S., Vargas, P. A.,and Braun, T.
    \newblock {The use of unmanned aerial vehicles and wireless sensor networks for spraying pesticides.}
    \newblock {\em Journal of Systems Architecture}, 60, 393-404, 2014.


    \bibitem{mogili}
    Mogili, UM. R. and  Deepak,B. B. V. L.
     \newblock {Review on Application of Drone Systems in Precision Agriculture}
      \newblock {\em Procedia Computer Science-RoSMa2018}, 133, 502-509, 2018.

     \bibitem{schwartz}
Schwartz, M. D.
     \newblock {Lecture 17: color}
     \newblock {\em Harvard University}.

    \bibitem{hsiehroy}
 Hsieh Fushing. and Roy T.
 \newblock {Complexity of Possibly-gapped Histogram and Analysis of Histogram
  (ANOHT)}.
 \newblock {\em Royal Socity-Open Science}, 2018.

  \bibitem{hsiehshanyu}
 Fushing H., Liu S.-Y., Hsieh Y.-C., and McCowan B.
 \newblock {From patterned response dependency to structured covariate
  dependency: categorical-pattern-matching}.
 \newblock {\em PLoS One}, 2018.



\bibitem{hsiehwang}
Fushing H. and Wang X.
 \newblock {Coarse- and fine-scale geometric information content of Multiclass Classification and implied Data-driven Intelligence.}
 \newblock {\em MLDM }, 2020.

 \bibitem{hsiehchou}
  Hsieh, Fushing and E. P. Chou.
 \newblock {Categorical Exploratory Data Analysis: From Multiclass Classification and Response Manifold Analytics perspectives of baseball pitching dynamics}.
 \newblock {\em arXiv: }2006.14411, 2020.

 \bibitem{hsiehturnbull}
 Hsieh, Fushing; Turnbull, Bruce W.
 \newblock{Nonparametric and semiparametric estimation of the receiver operating characteristic curve.}
 \newblock{\em Ann. Statist}, 1996


\end{thebibliography}

\end{document}